\documentclass[lettersize,journal]{IEEEtran}
\usepackage{amsmath,amsfonts}
\usepackage{algorithmic}
\usepackage{algorithm}
\usepackage{array}
\usepackage[caption=false,font=normalsize,labelfont=sf,textfont=sf]{subfig}
\usepackage{textcomp}
\usepackage{stfloats}
\usepackage{url}
\usepackage{verbatim}
\usepackage{graphicx}
\usepackage{cite}
\usepackage{booktabs}
\usepackage{multirow}
\usepackage{color}
\usepackage{makecell}
\usepackage{array}
\usepackage[dvipsnames]{xcolor}

\usepackage{amsmath, amssymb}

\usepackage{hyperref}

\begin{document}
\title{Decoupled Travel Planning with Behavior Forest}
\author{Duanyang Yuan\textsuperscript{\textdagger}, Sihang Zhou\textsuperscript{\textdagger}, Yanning Hou, Xiaoshu Chen, Haoyuan Chen, Ke Liang, Jiyuan Liu, Chuan Ma, $\text{Xinwang Liu}^*$, \textit{Senior Member, IEEE}, $\text{Jian Huang}^*$
\thanks{Duanyang Yuan, Sihang Zhou, Yanning Hou, Haoyuan Chen, Jian Huang are with the College of Intelligence Science and Technology, National University of Defense Technology, Changsha 410073, China. (e-mail: huang\_jian@nudt.edu.cn) \par
Xiaoshu Chen, Ke Liang and Xinwang Liu are with the College of Computer Science and Technology, National University of Defense Technology, Changsha 410073, China. (e-mail: xinwangliu@nudt.edu.cn)  \par
Jiyuan Liu is with the College of Systems Engineering, National University of Defense Technology, Changsha 410073, China. \par
Chuan Ma is with the College of Computer Science, Chongqing University, Chongqing 400044, China. \par
\textsuperscript{$*$} Corresponding Authors \par \textsuperscript{\textdagger}These authors contributed equally.
}
}

\markboth{arXiv}%
{Shell \MakeLowercase{\textit{et al.}}: A Sample Article Using IEEEtran.cls for IEEE Journals}


\maketitle

\begin{abstract}
Behavior sequences, composed of executable steps, serve as the operational foundation for multi-constraint planning problems such as travel planning.
In such tasks, each planning step is not only constrained locally but also influenced by global constraints spanning multiple subtasks, leading to a tightly coupled and complex decision process.
Existing travel planning methods typically rely on a single decision space that entangles all subtasks and constraints, failing to distinguish between locally acting constraints within a subtask and global constraints that span multiple subtasks. Consequently, the model is forced to jointly reason over local and global constraints at each decision step, increasing the reasoning burden and reducing planning efficiency.
To address this problem, we propose the Behavior Forest method. Specifically, our approach structures the decision-making process into a forest of parallel behavior trees, where each behavior tree is responsible for a subtask. A global coordination mechanism is introduced to orchestrate the interactions among these trees, enabling modular and coherent travel planning.
Within this framework, large language models (LLMs) are embedded as decision engines within behavior tree nodes, performing localized reasoning conditioned on task-specific constraints to generate candidate subplans and adapt decisions based on coordination feedback. The behavior trees, in turn, provide an explicit control structure that guides LLM generation.
This design decouples complex tasks and constraints into manageable subspaces, enabling task-specific reasoning and reducing the cognitive load of LLM.
Experimental results show that our method outperforms state-of-the-art methods by 6.67\% on the TravelPlanner benchmark and by 11.82\% at medium-level difficulty on the ChinaTravel benchmark, demonstrating its effectiveness in increasing LLM performance for complex multi-constraint travel planning \footnote{The code will be released after the paper is published}.
\end{abstract}



\section{Introduction}

\begin{figure}[t] 
    \centering
    \includegraphics[scale=1]{intro.png} 
    \caption{Illustration of the Behavior Forest framework. Behavior Forest organizes complex multi-constraint problems as a forest of parallel behavior trees that are connected through a global coordination mechanism. This framework decouples the task into multiple subtasks and disentangles the constraints into local and global constraints. Each tree in the forest handles a specific subtask, reasoning independently over its local constraints. This reasoning interacts with the LLM to generate executable plans, while global constraints are coordinated across all subtasks.}
    \label{intro1} 
\end{figure}

\IEEEPARstart{B}{ehavior} sequences transform vague instructions into clear and executable steps, providing the operational backbone for complex multi-constraint planning problems. Among such tasks, multi-constraint travel planning \cite{TravelPlanner} is especially challenging, as it involves multiple subtasks and interdependent constraints. While each constraint is specific, satisfying one often affects the feasibility options in others. For example, selecting a hotel room depends on the budget, which in turn is influenced by prior expenses such as transportation and dining. As a result, generating a valid plan within a single decision space is challenging, since all the constraints must be resolved simultaneously. Multi-step reasoning addresses this by progressively managing these dependencies and structuring decisions into coherent behavioral sequences.

Existing behavior sequence generation approaches can be generally categorized into traditional and LLM-based approaches. Traditional approaches first formalize natural-language problems as constraint satisfaction problems (CSPs) \cite{constraint-based, Heuristics, Incremental}, such as Boolean satisfiability (SAT) or satisfiability modulo theories (SMT), and then rely on algorithm-based solvers \cite{de2008z3, dutertre2006fast} to generate feasible plans. However, these methods usually rely on precise and complete constraint modeling and require iterative reformulation to handle complex multi-constraint planning, which limits their practicality \cite{RealWorld}. The rapid advancement of LLMs has reshaped this workflow \cite{Multi-ConstraintsCollaborative, DPPM, Plan-and-Solve}. LLMs can directly interpret natural-language inputs and generate plans \cite{guihypertree, gu2025agentgroupchat}, significantly simplifying the pipeline. Mechanisms \cite{liclosed, llm-modulo, kambhampati2024position, guo2025mirror, yuan2025evoagent}, such as ReAct \cite{ReAct} and Reflection \cite{Reflexion}, further improve the quality of the plan by interleaving reasoning with action and allowing iterative self-reflection based on feedback, respectively. 
Recent work further explores the use of intermediate representations in LLM-based planning, such as PDDL \cite{liu2023llm+}, signal temporal logic \cite{chen2024autotamp}, executable semantic objects \cite{chang2024lgmcts}, or executable code \cite{hsiao2025critical, RealWorld}, from which the final plan is derived. These approaches translate natural-language planning problems into verifiable symbolic or programmatic structures, thereby improving planning accuracy.
As research has advances, tree-based structures have become central to guiding complex reasoning \cite{biforest, gu2025agentgroupchat}. Tree-of-Thought (ToT) \cite{yao2023tree} models planning as a search over a tree, improving completeness and consistency. Following this approach, Hybrid ToT \cite{hu2023tree} combines fast and slow reasoning modes to strike a balance between planning efficiency and robustness. The HTP framework \cite{guihypertree} further extends the capabilities of Chain-of-Thought (CoT)\cite{chen2025putting} and ToT\cite{yao2023tree} by incorporating external rules, allowing LLMs to decompose complex tasks through a hypertree and to generate plans hierarchically.

Although recent LLM-based methods have demonstrated strong capabilities, they still suffer from fundamental limitations when handling complex multi-constraint planning problems. First, most existing approaches treat planning as a one-shot or implicitly decomposed process, where the LLM is required to generate a complete multi-day plan directly or to internally decompose the task. Consequently, multiple subtasks are presented to the LLM simultaneously, including transportation, accommodation, attractions, and dining. This forces the model to reason within an undifferentiated task space, thereby leading to planning failures. Second, current methods generally fail to distinguish between local constraints that are specific to individual subtasks and global constraints that coordinate multiple subtasks. These local and global constraints are entangled within the same LLM decision space and are injected into the reasoning process at each planning step. As a result, the LLM must consider all the constraints simultaneously, without mechanisms to isolate locally relevant constraints or to maintain global consistency in a structured manner. Third, planning is typically performed step-by-step, where each action relies on increasingly accumulated context embeddings. This design not only prevents parallel reasoning across subtasks but also lengthens the reasoning chain and amplifies error propagation, as mistakes made in early actions can persist and affect subsequent decisions. Even tree-based reasoning methods, while improving search completeness, often require the entire structured task information and constraints to be fed into the LLM at once, further overloading its reasoning capacity. Overall, these limitations reveal a fundamental bottleneck in existing LLM-based planning methods: tasks and constraints are tightly coupled within a single and entangled decision space. This coupling substantially increases the cognitive burden on the LLM, leading to missed constraints and repeated corrective actions. As the complexity of the problem increases, these limitations gradually degrade both planning efficiency and execution accuracy.

To solve this problem, we propose Behavior Forest, which structures multi-constraint planning as a forest structure of parallel trees connected through a global coordination mechanism. As shown in Figure \ref{intro1}, Behavior Forest decouples a complex task into multiple manageable subtasks and disentangles the constraints into local and global constraints. Each subtask is handled by a dedicated behavior tree that manages its local constraints, while global constraints coordinate the trees to ensure consistency. Candidate solutions that satisfy the constraints are collected through LLM interaction, while violations trigger subtask-level backtracking for coordination. This approach enables the forest to reason concurrently while remaining coordinated. The main contributions of this work are as follows:

\begin{itemize}
  \item We propose Behavior Forest, a novel paradigm that structures multi-constraint planning as a forest of parallel behavior trees. This design transforms sequential reasoning into modular and coherent subtask reasoning, while maintaining global consistency through a dedicated coordination mechanism, thereby improving overall reasoning efficiency.
  \item Each subtask is encapsulated within a behavior tree that governs its local constraints. This decoupling of tasks and constraints enables task-specific reasoning, reduces the cognitive load on LLMs, and mitigates error propagation across subtasks.
  \item Experimental results show that Behavior Forest outperforms state-of-the-art methods by 6.67\% on the TravelPlanner benchmark and by 11.82\% at the level of medium difficulty on the ChinaTravel benchmark.
\end{itemize}

\section{Related Work}
\subsection{Planning Methods}
Planning aims to generate action sequences that satisfy multiple constraints and achieve specified goals. To address action sequence generation in multi-constraint complex planning, numerous approaches have been proposed. These approaches differ primarily in how each planning decision is generated. We categorize existing planning methods into two types: solver-based and LLM-based approaches.
\subsubsection{Solver-based Planning Methods}
To address multi-constraint planning problems such as travel planning, solver-based approaches typically formalize the task as a constraint satisfaction problem (CSP) \cite{constraint-based}, including formulations based on Boolean satisfiability (SAT) \cite{Heuristics} and satisfiability modulo theories (SMT) \cite{Incremental}. These formulations are then solved using algorithm-based solvers \cite{dutertre2006fast, de2008z3}. For example, the SMT framework \cite{llm-modulo} combined with iterative back-prompting is employed to adjust the user input and correct errors. The LLM-based tools are then used for itinerary generation, parsing, and feedback, allowing the prompts to be refined iteratively. The llm-rwplanning \cite{RealWorld} leverages LLMs for prompt processing and employs non-LLM SMT solvers together with multiple APIs (CitySearch, FlightSearch, AttractionSearch, DistanceSearch, AccommodationSearch, and RestaurantSearch) to collect structured information for itinerary planning. However, solver-based methods face several challenges, including steep learning curves and limited ability to extract information from natural-language input accurately. Even when information is extracted correctly, users often need to revise their inputs and perform multiple iterations to resolve unsatisfiable queries \cite{RealWorld}.

\subsubsection{LLM-based Planning Methods}
Recent work has demonstrated that LLMs achieve promising performance on planning tasks, benefiting from their rich world knowledge, strong reasoning abilities, and tool-use capabilities\cite{zhang2025collm, xu2024global, few-shotlearners}.
Existing LLM-based planning methods can be categorized into two types: plan-then-execute and step-by-step approaches. Plan-then-execute methods \cite{Multi-ConstraintsCollaborative, DPPM, guihypertree, Incontext1, Incontext2, wang2025large, Hugginggpt, Plan-and-Solve} first generate a complete plan for each step or subtask and then execute it. Although simple, they are susceptible to error propagation and uneven constraint allocation, often exceeding the reasoning capacity of LLMs.
Step-by-step approaches \cite{liclosed, kambhampati2024position, Chain-of-thought, Pal, liu2024learning} alternate between task decomposition and planning while incorporating self-reflection. Although these iterative strategies reduce error accumulation, each iteration addresses only local constraints and lacks a global perspective, which can lead to subtask conflicts and incoherent overall planning. Moreover, repeated cycles introduce significant computational overhead, which reduces planning efficiency.
Hybrid methods that combine retrieval-augmented generation (RAG) \cite{verma2025leap} offer external knowledge support, helping to alleviate some conflicts. However, they remain susceptible to reliability issues from noisy retrieval, which can compromise action selection and lead to inaccurate execution.

\subsection{Behavior Tree}

Behavior trees structure the planning process by hierarchically decomposing tasks, alleviating the burden of considering all the constraints simultaneously. However, traditional behavior trees are based on manually designed structures and predefined rules, which limit their flexibility and scalability. To address this, LLM-enhanced behavior trees leverage the reasoning and knowledge capabilities of LLMs to dynamically generate tree structures. These LLM-enhanced behavior trees can be classified into three categories: direct generation \cite{Multi-Robot,Free-Form,Mrbtp}, reflection-based \cite{aaai, LLM-Driven,Hbtp}, and state-based generation methods \cite{StarCraftII,manipulationICRA}. 
The first type refers to LLMs that generate behavior trees in a single step, accounting for all tasks at once. However, these approaches struggle with large-scale reasoning, often producing trees that violate constraints. The second type employs reflection mechanisms, where LLMs first generate draft trees and iteratively refine them through self-reflection. This approach involves frequent backtracking, which significantly increases the computational cost. The third type adopts state-based generation that involves constructing the behavior tree one action at a time. Although effective in simplifying each decision, these serial paradigms inherently limit the model’s ability to anticipate downstream consequences. These methods lack parallel consideration of interdependent subtasks and still require reasoning over the full constraint space at each decision point.

\subsection{LLM-powered Multi-Agent Systems}
Single-agent systems suffer from hallucinations and limited contextual capacity, rendering them inadequate for complex real-world tasks.
To overcome these issues, researchers have developed multi-agent systems \cite{cai2025survey, sun2024optimizing}. These systems can be categorized into three main types: decomposition-based approaches \cite{cheng2025evocurr,DPPM, yidebate}, hierarchy-based approaches \cite{wang2025megaagent}, and discussion-based approaches \cite{Tree-of-Reasoning,chen2025mediator}.
In decomposition-based approaches, each agent is assigned a subtask, and the results are subsequently integrated. For example, EvoCurr \cite{cheng2025evocurr} employs a curriculum-learning framework. In this framework, a curriculum-generation agent guides a code-generation agent to construct decision trees from simple to complex tasks, improving specialization but requiring a time-consuming training process. Similarly, DPPM \cite{DPPM} divides travel planning into transportation, accommodation, dining, and attraction planning. Each agent handles the constraints relevant to its subtask, but the method lacks global planning across subtasks. The final plan is generated by taking the Cartesian product of all the subtasks, which greatly increases the computation. In addition, as the method operates in a sole-planning stage without considering multiple API calls, it is difficult to apply in real-world scenarios. 
Hierarchy-based approaches, such as MegaAgent \cite{wang2025megaagent}, organize task execution through hierarchical decomposition. The BossAgent coordinates the process by assigning tasks to the AdminAgents, each capable of further delegating responsibilities to subordinate agents. 
On the other hand, discussion-based approaches focus on cooperation and deliberation among agents. Systems such as Tree-of-Reasoning \cite{Tree-of-Reasoning} and mediator-guided collaboration \cite{chen2025mediator} assign function-specific agents to exchange, refine, and reconcile reasoning paths through dialog. Although this improves reasoning diversity, the results may still be unstable because of inconsistent LLM judgment, and excessive collaboration can paradoxically shift correct answers into incorrect ones.

\section{Method}
\begin{figure*}[t] 
    \centering
    \includegraphics[scale=0.8]{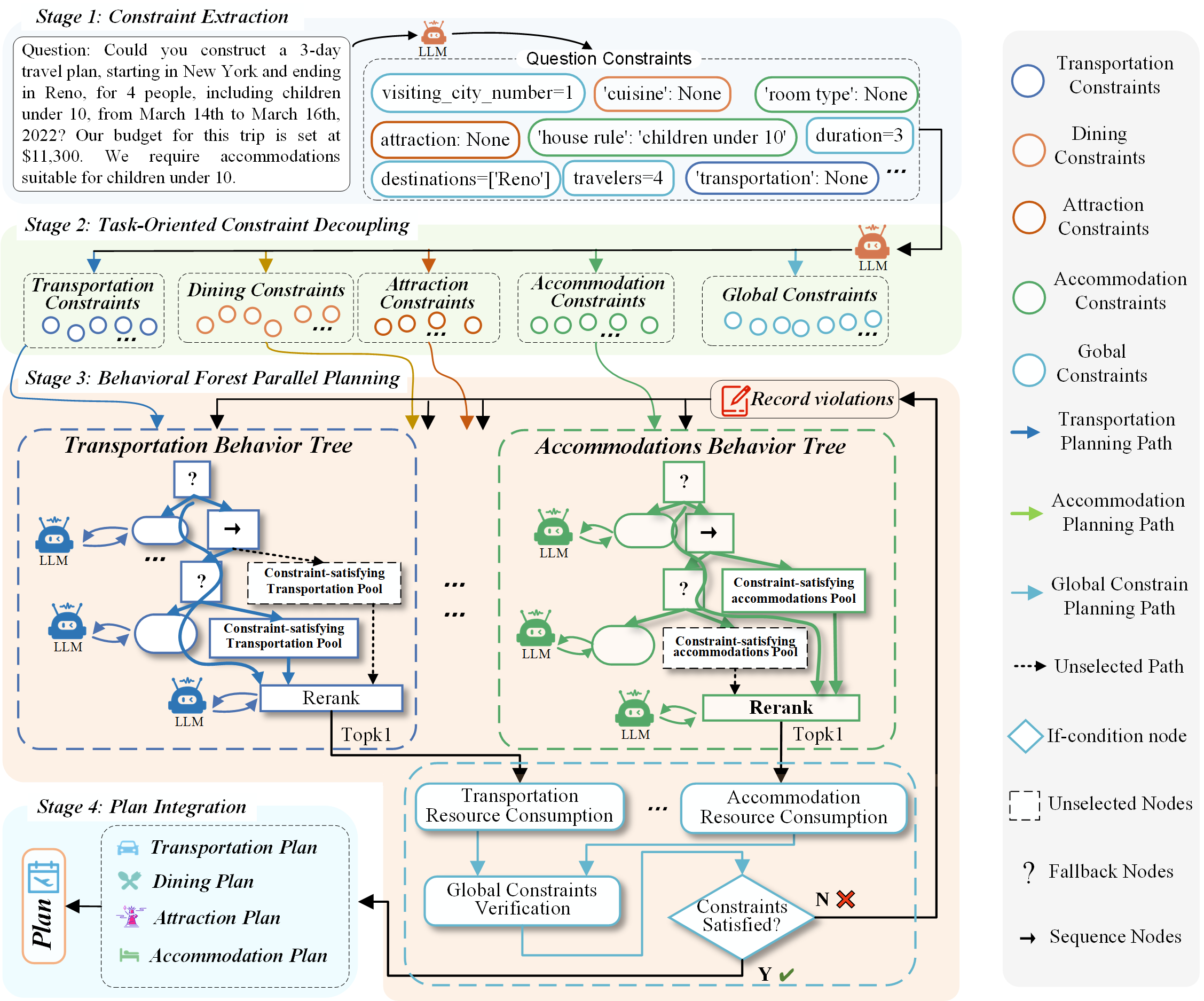} 
    \caption{Overall framework of Behavior Forest. Stage~1 extracts constraint information from natural-language queries using LLMs, such as dates, budgets, user preferences, and other relevant information. Stage~2 decouples the task into four subtasks—transportation, dining, attractions, and accommodation—and disentangles the extracted constraints into local and global constraints. Local constraints are assigned to the subspaces corresponding to the four subtasks. Stage~3 performs parallel planning, where each tree handles one subtask. Each tree interacts with the LLM to generate a constraint-satisfying candidate pool, whose elements are then reranked using heuristic scoring to produce feasible solutions. Global constraints connect the trees to ensure cross-tree consistency, such as by verifying budgets in travel planning, and recording failed plan combinations to avoid regenerating the same infeasible solutions. Finally, Stage~4 integrates the feasible subplans into a complete plan.}
    \label{method}
\end{figure*}

\subsection{Formalization of Behavior Forest}
Given a natural-language travel query $Q$, Behavior Forest formulates planning as parallel behavior trees connected through a global coordination mechanism, with the overall procedure consisting of the following steps.

\subsubsection{Constraint Extraction}
Before planning, the constraint information embedded in $Q$ is explicitly extracted using an LLM. Specifically, the LLM parses the natural-language query into a structured constraint set:
\begin{equation}
    \psi = \text{LLM-Parser}(Q) \;\Rightarrow\; C,
\end{equation}
where $C$ denotes the set of constraints implied by the query, including both explicit requirements and implicit preferences. $\text{LLM-Parser}(\cdot)$ denotes the process of LLM parsing the natural-language query. This transformation enables the system to convert unstructured natural language into a structured constraint set.

\subsubsection{Task-Oriented Constraint Decoupling}
After the constraints are extracted from the query, the planning problem is further decoupled into these tasks and their associated constraints.

\textbf{Task Decoupling.}
An LLM is used to decouple the natural-language travel query $Q$ into a set of planning tasks:
\begin{equation}
\mathcal{D}(Q) = \{\tau_1, \tau_2, \dots, \tau_m\},
\end{equation}
where $\mathcal{D}(Q)$ denotes the decoupled task set. Each task $\tau_i$ corresponds to an independent decision dimension. 

Then, for the decoupled task set $\mathcal{D}(Q)$, tasks are assign to each behavior tree as follows:
\begin{equation}
\mathcal{F}(Q) = \{\mathcal{T}_1, \mathcal{T}_2, \dots, \mathcal{T}_m\},
\end{equation}
where each behavior tree $\mathcal{T}_i$ is responsible for planning task $\tau_i$. $\mathcal{F}(Q)$ represents the set of behavior trees $\mathcal{T}_i$ for query $Q$.

\textbf{Constraint Decoupling.}
Given the extracted constraint set $C$ from the query $Q$, the framework further partitions $C$ into task-specific local constraints and global constraints.
Formally, the constraint set is decoupled as follows:
\begin{equation}
C = \Big( \bigcup_{i=1}^{m} C_i \Big) \cup C^{\mathrm{global}},
\end{equation}
where each subset $C_i$ denotes the set of local constraints relevant only to task $\tau_i$, and $C^{\mathrm{global}}$ represents global constraints that involve cross-task dependencies. This decoupling allows each behavior tree $\mathcal{T}_i$ to focus on the reasoning subspace defined by its constraint subset $C_i$, isolating irrelevant factors and reducing cognitive entanglement. By partitioning the constraint space, Behavior Forest enables parallelized planning, where trees operate concurrently yet remain interconnected through the global coordination constraint.

Each behavior tree is then defined as a tuple:
\begin{equation}
\mathcal{T}_i = \langle N_i, E_i, C_i \rangle,
\end{equation}
where $N_i$ denotes the set of decision nodes associated with task $\tau_i$,
$E_i \subseteq N_i \times N_i$ represents the hierarchical control-flow structure,
and $C_i \subseteq C$ is the set of local constraints assigned to task $\tau_i$.
This constraint assignment is performed after task decoupling, ensuring that each tree reasons only over task-relevant constraints, while global constraints are enforced through the coordination mechanism.

\subsubsection{Behavior Forest Parallel Planning}
Each behavior tree $\mathcal{T}_i$ independently performs planning for its assigned task under its associated constraint set $C_i$.
The planning result of each tree is a local subplan:
\begin{equation}
    plan_i = \mathrm{Plan}(\mathcal{T}_i \mid C_i),
\end{equation}
where $\mathrm{Plan}(\cdot)$ denotes the behavior tree-based planning process guided by the LLM, and $plan_i$ represents the complete planning result produced by $\mathcal{T}_i$.

During planning, the behavior tree does not reason over all nodes or constraints simultaneously.
Instead, at each node expansion or decision step, the LLM conditions its reasoning on an activated subset of constraints:
\begin{equation}
    C_i^{\mathrm{act}} \subseteq C_i,
\end{equation}
where $C_i^{\mathrm{act}}$ contains only the constraints relevant to the current node and planning stage of $\mathcal{T}_i$. This design bounds LLM reasoning to a task-specific decision subspace and mitigates cross-task interference.

Given the set of subplans $\{plan_1, plan_2, \dots, plan_m\}$, Behavior Forest applies a global coordination mechanism $C^{\mathrm{global}}$ to iteratively validate their joint feasibility. 
If a combination of subplans violates any constraint in $C^{\mathrm{global}}$, the corresponding local plans $\{plan_i\}$ are recorded as infeasible to prevent the same invalid combination from regenerating.
Moreover, the violation signal is propagated back to the associated behavior trees, triggering plan updates and the generation of new candidate subplans.
This iterative process continues until either a feasible set of subplans satisfying all global constraints is found or a maximum number of coordination iterations is reached.

\subsubsection{Plan Integration}
After each behavior tree has produced its subplan, these subplans are subsequently integrated to form the final travel plan:
\begin{equation}
Plan = \mathrm{Integrate}\big(\{plan_1, plan_2, \dots, plan_m\}\big),
\end{equation}
where $\mathrm{Integrate}(\cdot)$ assembles these subplans into a coherent end-to-end itinerary. $\textit{Plan}$ denotes the final plan composed of the local plans. 

\subsection{Travel Planning with Behavior Forest}

We formulate travel planning as a multi-task and multi-constraint planning problem grounded in a natural-language query. Given a user query $Q$, the goal is to generate a feasible and coherent travel plan $Plan$ that satisfies user-specified constraints. Formally, the problem is defined as follows:
\begin{equation}
    P = \langle Q, C, R, O \rangle,
\end{equation}
where each component is defined as follows.

\textbf{Query \(Q\).}  
Represents a natural-language travel request expressed by the user.
The query $Q$ is provided in text and implicitly contains information such as the origin city, candidate destinations, travel duration, budget limit, number of travelers, user preferences, and other constraints.
In addition, the overall travel plan consists of multiple interrelated subplans.
Accordingly, the task decoupling $\mathcal{D}(Q) = \{\tau_1, \tau_2, \dots, \tau_m\}$ instantiates these subplans as task-specific planning dimensions, where $\tau_i$ corresponds to transportation, accommodation, dining, or attractions, respectively.

\textbf{Constraint Set $C$.}   
$C = \{c_1, c_2, \dots, c_n\}$ denotes the set of constraints extracted from $Q$, including hard constraints (e.g., budget) and soft constraints (e.g., preferences).  
$C$ is decoupled into local constraints $\{C_i\}$ and global constraints $C^{\mathrm{global}}$.
$C^{\mathrm{global}}$ captures cross-task dependencies, such as budget limits. The local constraints $\{C_i\}$  are assigned to multiple trees in $\mathcal{F}(Q)$.

\textbf{Resource Set $\mathcal{R}$.}  
The planning space is grounded in domain-specific resources:
\begin{align}
\begin{split}
    \mathcal{R} = \{ 
    &R_T: \text{transportation options}, \\
    &R_A: \text{accommodation options}, \\
    &R_D: \text{dining options}, \\
    &R_{AT}: \text{attraction options}
    \}.
\end{split}
\end{align}

\textbf{Objective Function $O$.}
In Behavior Forest, multiple behavior trees are planned in parallel over different decision dimensions while sharing a global budget constraint.
As each tree $\mathcal{T}_i$ generates its local plan $\textit{plan}_i$, it simultaneously contributes to the overall budget consumption, forming a shared budget state across the trees. Formally, let $B$ denote the total budget extracted from the query $Q$. The cost variation induced by a local plan $\textit{plan}_i$ is represented as follows:
\begin{equation}
    \Delta \textit{Cost}(\textit{plan}_i, B_i),
\end{equation}
where $B_i$ denotes the budget allocation associated with tree $\mathcal{T}_i$. $\Delta \textit{Cost}(\cdot)$ captures the budget consumption of $\mathcal{T}_i$ under its current planning decision.

These cost variations from all the trees are aggregated and synchronously coordinated through the global coordination structure $C^{\mathrm{global}}$:
\begin{equation}
    \Delta \textit{Cost}(\textit{plan}_i, B_i)
    \;\Rightarrow\;
    \textit{Update}(B_j),
    \quad
    \forall \langle \mathcal{T}_i, \mathcal{T}_j, \omega_{ij} \rangle \in C^{\mathrm{global}}.
\end{equation}
where $B_j$ denotes the budget allocation associated with tree $\mathcal{T}_j$. $\textit{Update}(B_j)$ represents a synchronous update operation applied to the budget state of another behavior tree $\mathcal{T}_j$, reflecting the impact of the decision of $\mathcal{T}_i$ on the feasible budget space of $\mathcal{T}_j$. The $\langle \mathcal{T}_i, \mathcal{T}_j, \omega_{ij} \rangle$ specifies a dependency relation between two behavior trees.

This budget feedback mechanism enforces global budget consistency by jointly constraining the decisions of all behavior trees during parallel planning.
In this way, the budget serves as a shared coordination variable that connects independent behavior trees, enabling coordinated reasoning and mutual constraint satisfaction across the trees.

The objective of this coordinated planning process is to obtain a globally feasible travel plan $\textit{Plan}$ that satisfies all extracted constraints $C$ in the given budget $B$.
This objective is defined as follows:
\begin{equation}
    O = \text{LLM}(\textit{prompt}, \textit{Plan}, C),
\end{equation}
where $\textit{Plan}$ denotes the final plan composed of the local plans,
$\textit{prompt}$ represents the instruction template provided to the LLM,
and $C$ denotes the entire constraint set derived from the natural-language query.

Through iterative parallel planning, synchronous budget feedback, and global coordination, Behavior Forest progressively refines candidate plans toward optimizing the objective $O$ while maintaining global feasibility.

\section{Experiments}
\subsection{Experimental Settings}
\subsubsection{Benchmarks}
To evaluate the performance of Behavior Forest in multi-constraint travel planning scenarios, we conduct experiments on the TravelPlanner \cite{TravelPlanner} and ChinaTravel \cite{shao2024chinatravel} datasets. For the TravelPlanner dataset, we follow both the sole-planning and two-stage settings. In the sole-planning setting, LLMs generate plans based on the information provided, without the need for external tools such as databases to retrieve flight departure times. For the ChinaTravel dataset, we conduct experiments at three difficulty levels: easy, medium, and human. Both datasets include diverse travel queries with varying travel durations and complexity.
Given user-specified origins, destinations, and personal requirements, agents are tasked with generating travel plans that satisfy multiple constraints.

\subsubsection{Baselines}
For the TravelPlanner dataset, we compare Behavior Forest with the following two-stage baselines: ReAct \cite{ReAct}, EvoAgent \cite{yuan2025evoagent}, Human-Like \cite{xie2024human}, Co-Gym \cite{shao2024collaborative}, Optimizing \cite{wei2024optimizing}, LLM-Modulo \cite{llm-modulo, kambhampati2024position}, PMC \cite{Multi-ConstraintsCollaborative}, RLP \cite{xie2024revealing}, FAFT \cite{chen2024can}, MIRROR \cite{guo2025mirror}, NarrativeGuide \cite{zhang2025narrative}, HTP \cite{guihypertree}, ToT \cite{yao2023tree}, Lats \cite{zhou2024language}, One-shot learning, HiAR-ICL \cite{wu2024beyond}, RAP \cite{hao2023reasoning}, and llm-rwplanning \cite{RealWorld}. We also compare the following sole-planning baselines: (1) Direct planning method \cite{TravelPlanner} uses LLMs to directly generate travel plans based on the provided reference information. (2) CoT \cite{wei2022chain} uses chain-of-thought instructions to guide intermediate reasoning steps, enabling systematic task decomposition and planning. (3) ReAct \cite{ReAct} applies a reasoning-and-action framework for each decision. (4) Reflection \cite{Reflexion} employs a reasoning-action-reflection framework for decision-making. (5) EvoAgent \cite{yuan2025evoagent} leverages evolutionary algorithms to scale specialized agents into multi-agent systems. (6) DPPM \cite{DPPM} decomposes complex tasks into subtasks based on constraints.

For the ChinaTravel dataset, we compare Behavior Forest with the following baselines: (1) ReAct \cite{ReAct}, a widely-adopted reasoning-and-action framework, along with its act-only ablation variant. (2) NeSy Planning \cite{pan2023logic}, which integrates LLMs with symbolic solvers to improve logical problem-solving. (3) LLM-Modulo \cite{llm-modulo, kambhampati2024position}, which combines LLM with external verifiers through a tightly coupled bi-directional interaction.

\subsubsection{Implementation Details}
In our experiments, we limit the maximum number of retries to 3 to prevent infinite remaking of the travel plan. We evaluate our approach using DeepSeek-V3, GPT-3.5, and Mistral-7B, and run the open-source models on 2×4090-24G GPUs. The temperature is fixed at 0.7 for all processes.

\subsubsection{Evaluation Metrics}
We use the same metrics and settings as proposed in TravelPlanner \cite{TravelPlanner}.

Delivery Rate: This metric evaluates the agent's ability to successfully complete a plan within a number of steps. Failure occurs if the agent gets stuck in a loop, encounters multiple unsuccessful attempts, or exceeds the step limit. 

Commonsense Pass Rate: This metric examines the agent’s ability to integrate commonsense knowledge into the plan without receiving explicit guidance. It consists of eight distinct commonsense dimensions. 

Hard Pass Rate: This metric assesses whether a plan adheres to all the explicitly defined hard constraints in the given query, testing the agent's ability to reliably adapt plans to diverse and varying user requirements. 

Final Pass Rate: This metric reflects the percentage of feasible plans that meet all the aforementioned constraints, indicating the agent's proficiency in generating plans that adhere to practical standards. 

Micro Pass Rate: This metric measures the proportion of constraints successfully satisfied across all plans. It is calculated by dividing the total number of constraints met by the total number of constraints applied to the plans. The micro pass rate is as shown in Formula \eqref{MIPR}.

\begin{equation}
{\text{Micro Pass Rate}=\frac{\sum_{p\in P}\sum_{c\in C_p}1_{\mathrm{passed}(c,p)}}{\sum_{p\in P}|C_p|}}.
\label{MIPR}
\end{equation}

Macro Pass Rate: This metric gauges the percentage of plans that fulfill all their constraints. It is calculated by dividing the number of plans that satisfy all relevant commonsense or hard constraints by the total number of plans tested. The macro pass rate is as shown in formula \eqref{MAPR}.

\begin{equation}
{\text{Macro Pass Rate}=\frac{\sum_{p\in P}1_{\mathrm{passed}(C_p,p)}}{|P|}}.
\label{MAPR}
\end{equation}

\subsection{Main Results}
\subsubsection{Results on the TravelPlanner Dataset}

\begin{table*}[]
\centering
\caption{Comparison of the performance of state-of-the-art methods on the validation set of the TravelPlanner benchmark. $*$ indicates that the results of the corresponding method are reproduced by our work under the same experimental settings. The best scores in each column are highlighted in \textbf{bold}, while the second-best scores are \underline{underlined}. Overall, the Behavior Forest method achieves the highest performance across all metrics.}
\scalebox{1.2}{
\begin{tabular}{ccccccc} \hline
\multirow{2}{*}{Method} & \multirow{2}{*}{Delivery Rate} &
\multicolumn{2}{c}{Commonsense Pass Rate} & \multicolumn{2}{c}{Hard Pass Rate} &
\multirow{2}{*}{Final Pass Rate}\\ \cmidrule(lr){3-4}\cmidrule(lr){5-6}
 &  & Micro & Macro & Micro & Macro & \\ \hline
\multicolumn{7}{c}{\textbf{Two-stage}} \\ \hline
ReAct\textsuperscript{*}\cite{ReAct}   & 85.00  & 53.40    & 2.78   & 2.14    & 2.78   & 1.11     \\ 
MIRROR\cite{guo2025mirror}               & \textbf{100.00} & 73.20 & 13.90 & 27.60 & 13.30 & 2.20 \\
Human-Like\cite{xie2024human}   & \textbf{100.00} & 75.10   & 15.60  & 15.50  & 4.40      & 2.20   \\
ToT\cite{yao2023tree}                  & \textbf{100.00} & 75.80 & 16.70 & 44.00 & 21.10 & 3.90 \\
Lats\cite{zhou2024language}                 & \textbf{100.00} & 78.30 & 16.70 & 44.80 & 21.10 & 3.90 \\
One-shot\cite{guihypertree}             & \textbf{100.00} & 76.40 & 17.20 & 43.30 & 20.60 & 3.90 \\
EvoAgent\textsuperscript{*}\cite{yuan2025evoagent}  & \textbf{100.00}    & 76.25  & 15.10 & 38.57   & 28.42    & 4.44     \\
HiAR-ICL\cite{wu2024beyond}             & \textbf{100.00} & 81.20 & 20.60 & 47.40 & 22.80 & 6.70 \\
RAP\cite{hao2023reasoning}                  & \textbf{100.00} & 81.00 & 20.60 & 50.50 & 25.60 & 6.70 \\
Co-Gym\cite{shao2024collaborative}        & 85.30 & - & - & - & - & - \\
Optimizing\cite{wei2024optimizing}           & \textbf{100.00} & 84.93 & 27.22 & 61.19 & 42.78 & 12.78 \\
FAFT\cite{chen2024can}         & \textbf{100.00} & 89.10 & 59.40 & 49.80 & 21.70 & 13.90 \\
HTP\cite{guihypertree}                  & \textbf{100.00} & 87.20 & 37.80 & 44.30 & 32.20 & 20.00 \\
LLM-Modulo\cite{llm-modulo, kambhampati2024position}      & \textbf{100.00} & 89.20 & 40.60 & 62.10 & 39.40 & 20.60 \\
RLP\cite{xie2024revealing}            & - & \underline{95.30} & 68.90 & 62.60 & 39.40 & 25.00 \\
PMC\textsuperscript{*}\cite{Multi-ConstraintsCollaborative}    & 97.33 & 86.11 & 49.44 & 49.76 & 50.00 & 30.00 \\
NarrativeGuide\cite{zhang2025narrative}       & \underline{99.40} & 94.30 & 63.90 & 64.00 & 43.30 & 34.40 \\
llm-rwplanning\textsuperscript{*}\cite{RealWorld} & 90.00 & 90.00 & \underline{90.00} &\underline{85.95} & \underline{85.00} & \underline{85.00} \\
\textbf{Behavior Forest} & \textbf{100.00} & \textbf{98.40} & \textbf{94.44} & \textbf{90.00} & \textbf{91.67} & \textbf{91.67} \\ \hline
\multicolumn{7}{c}{\textbf{Sole-planning}} \\ \hline
Cot\textsuperscript{*}\cite{wei2022chain}        & \textbf{100.00} & 63.33 & 0.00 & 0.00 & 0.00 & 0.00 \\
Direct\textsuperscript{*}\cite{TravelPlanner}     & \textbf{100.00} & 63.26 & 2.22 & 2.62 & 3.33 & 0.56 \\
Reflection\textsuperscript{*}\cite{Reflexion}  & 76.11 & 63.05 & 16.11 & 32.38 & 22.22 & 5.56 \\
ReAct\textsuperscript{*}\cite{ReAct}      & \underline{85.00} & 70.07 & 18.89 & 47.86 & 33.89 & 6.67 \\
EvoAgent\textsuperscript{*}\cite{yuan2025evoagent}   & \textbf{100.00} & 84.58 & 35.00 & 58.81 & 62.22 & 43.33 \\
DPPM\cite{DPPM}       & \textbf{100.00} & \underline{98.50} & \underline{89.40} & \underline{90.70} & \underline{80.60} & \underline{76.70} \\
\textbf{Behavior Forest} & \textbf{100.00} & \textbf{99.44} & \textbf{96.11} & \textbf{93.81} & \textbf{94.44} & \textbf{94.44} \\ \hline
\end{tabular}
}
\label{mainresult}
\end{table*}

The experimental results in Table \ref{mainresult} demonstrate the superior performance of Behavior Forest in both evaluation settings. In the two-stage scenario, Behavior Forest consistently outperforms all baselines, with a final pass rate of 91.67\%, surpassing the strongest prior method, llm-rwplanning, by 6.67\%. The model also improves both micro-level metrics, achieving higher commonsense and hard micro pass rates than llm-rwplanning by 8.40\% and 4.05\%, respectively. These results collectively demonstrate that Behavior Forest effectively handles complex interdependent constraints.  
The performance gain stems from the forest-based reasoning architecture. This design decouples constraints across tasks, allowing each behavior tree to operate under task-specific constraints while maintaining global coordination. This design prevents constraint interference among trees. Consequently, the LLM produces more coherent multi-stage plans, reducing redundant information and stabilizing the reasoning process across intermediate steps. Compared with serial reasoning frameworks such as ReAct, Behavior Forest yields final pass rates that are more than 90\% higher, confirming that explicit constraint decoupling and inter-tree coordination substantially increase planning efficiency.

In the sole-planning configuration, Behavior Forest achieves a final pass rate of 94.44\%, surpassing DPPM by 17.74\%. Compared with other methods such as EvoAgent or Reflection, our method demonstrates superior constraint satisfaction, achieving a 96.11\% commonsense macro pass rate and a 94.44\% hard macro pass rate. Methods such as CoT fail to generate valid plans under this multi-constraint problem, as the requirement for plan generation directly imposes a high reasoning burden on the model, resulting in plan generation failures. This improvement stems from the hierarchical behavior tree structure, which enables fine-grained constraint tracking and pruning, reducing cascading reasoning errors. Overall, the results confirm that Behavior Forest retains the efficiency and robust reasoning in the sole-planning setting, confirming the scalability of its constraint-oriented forest architecture for travel planning.

\subsubsection{Results on the ChinaTravel Dataset}

\begin{table*}[]
\centering
\caption{Comparison of the performance of state-of-the-art methods on the ChinaTravel dataset. $*$ indicates that the results of the corresponding method are reproduced by our work under the same experimental settings. The dataset is divided into three difficulty levels: Easy, Medium, and Human. The best performance in each column is highlighted in \textbf{bold}, while the second-best scores are \underline{underlined}. Across all levels, Behavior Forest consistently achieves the highest scores.}
\scalebox{1.2}{
\begin{tabular}{ccccccc} \hline   
\multirow{2}{*}{}    & \multirow{2}{*}{Delivery Rate} & \multicolumn{2}{c}{Commonsense Pass Rate}   & \multicolumn{2}{c}{Hard Pass Rate}          & \multirow{2}{*}{Final Pass Rate}  \\\cmidrule(lr){3-4}\cmidrule(lr){5-6}
     &      & Micro                & Macro                & Micro  & Macro      &       \\\hline   
\multicolumn{7}{c}{\textbf{Easy}}      \\\hline   
Act\textsuperscript{*}\cite{ReAct}                  & 86.00                             & 65.40   & 0.00                    & 65.70                 & 31.67                & 0.00            \\
ReAct\textsuperscript{*}\cite{ReAct}   & \underline{90.00}   & 57.71   & 1.33  & \underline{82.27}   & 49.00   & 1.33         \\
LLM-Modulo\textsuperscript{*}\cite{llm-modulo, kambhampati2024position}     & 56.33   & \underline{94.49}  & 3.33    & 65.29  & 43.67   & 3.00          \\
NeSy Planning\textsuperscript{*}\cite{pan2023logic} & 73.00 & 73.00  & \underline{73.00} & 70.23   & \underline{61.67}   & \underline{61.67}       \\
\textbf{Behavior Forest} &\textbf{98.89} &\textbf{95.61} &\textbf{88.30} &\textbf{98.28} &\textbf{84.95} &\textbf{84.95}     \\\hline   
\multicolumn{7}{c}{\textbf{Medium}}     \\\hline   
Act\textsuperscript{*}\cite{ReAct}                  & 71.10                           & 32.60                 & 0.00                    & 60.30      & 45.10                & 0.00           \\
ReAct\textsuperscript{*}\cite{ReAct}                & 65.33            & 53.95                & 0.67                 & 78.64                & 50.67                & 0.67        \\
LLM-Modulo\textsuperscript{*}\cite{llm-modulo, kambhampati2024position}   & 56.67  &\underline{92.21} & 3.33  & 61.08   & 22.67   & 1.33      \\
NeSy Planning\textsuperscript{*}\cite{pan2023logic}        & \underline{84.67}   & 84.67 & \underline{84.67}  & \underline{82.28}   & \underline{66.00}  & \underline{66.00}  \\
\textbf{Behavior Forest}         & \textbf{95.41}                          & \textbf{93.98}                & \textbf{87.23}                & \textbf{94.15}                & \textbf{77.82}                & \textbf{77.82}      \\\hline   
\multicolumn{7}{c}{\textbf{Human}}   \\\hline   
Act\textsuperscript{*}\cite{ReAct}                  & 35.34                          & 16.80                 & 0.00                    & 35.74                & 25.81      & 0.00       \\
ReAct\textsuperscript{*}\cite{ReAct}                & 42.21                          & 33.15                & 0.00                    & 48.91                 & 36.36 & 0.00       \\
LLM-Modulo\textsuperscript{*}\cite{llm-modulo, kambhampati2024position}  & \underline{68.83} & \underline{90.58}  & 1.3 & \underline{76.42} & \underline{45.45}   & 1.30     \\
NeSy Planning\textsuperscript{*}\cite{pan2023logic}        & 62.99       & 62.91  & \underline{61.04} & 54.30  & 41.56                & \underline{41.56}    \\
\textbf{Behavior Forest}         & \textbf{96.73}                          & \textbf{90.62}                & \textbf{79.97}                & \textbf{93.83}                & \textbf{68.84}                & \textbf{68.84}   \\ \hline
\end{tabular}
}
\label{mainresult2}
\end{table*}

The experimental results in Table \ref{mainresult2} across different difficulty levels demonstrate the superiority of our method over baseline methods. In the easy setting, our method achieves a final pass rate of 84.95\%, outperforming the other baselines, while maintaining high delivery rates above 95\%. As the task complexity increases, the final pass rates of our method decrease from 84.95\% to 77.82\% and 68.84\%. However, it consistently maintains significant performance over the SOTA methods. Moreover, our method demonstrates strong macro metrics across all difficulty levels, indicating robust capabilities regardless of task complexity.
Notably, both Act and ReAct exhibit near-zero macro scores in commonsense pass rates because the macro metric requires all types of constraints to be simultaneously satisfied. Each planning query involves multiple interdependent constraints, making it difficult for the models to satisfy all categories simultaneously. Consequently, macro scores remain low, particularly in the human setting. This discrepancy further highlights the robustness of our method in handling complex tasks.

\textit{\textbf{Summary}} On the TravelPlanner dataset, Behavior Forest achieves final pass rates of 91.67\% and 94.44\% under two-stage and sole-planning settings, respectively. On the ChinaTravel dataset, Behavior Forest achieves final pass rates of 84.95\%, 77.82\%, and 68.84\% at levels of easy, medium, and human difficulty, respectively. These results outperform those of all baseline methods, demonstrating the effectiveness of Behavior Forest.

\subsection{Performance Analysis Across Different Constraint Types}

\begin{table*}[]
\centering
\caption{Performance comparison across different methods under various commonsense and hard constraints. The best performance for each difficulty level is highlighted in \textbf{bold}, while the second-best scores are underlined using \underline{underline}. Behavior Forest achieves the highest scores across most constraint types.}
\scalebox{1.1}{
\begin{tabular}{cccccccccc} \hline
\multirow{2}{*}{Constraint Types} & \multicolumn{3}{c}{Behavior Forest}  & \multicolumn{3}{c}{PMC\cite{Multi-ConstraintsCollaborative}}   & \multicolumn{3}{c}{llm-rwplanning\cite{RealWorld}} \\ \cmidrule(lr){2-4}\cmidrule(lr){5-7}\cmidrule(lr){8-10}
  & Easy   & \multicolumn{1}{c}{Medium} & \multicolumn{1}{c}{Hard}  & Easy   & \multicolumn{1}{c}{Medium} & \multicolumn{1}{c}{Hard} & Easy  & \multicolumn{1}{c}{Medium} & \multicolumn{1}{c}{Hard}\\ \hline
\multicolumn{10}{c}{Commonsense Constraint}  \\ \hline
Within Sandbox      & \textbf{100.00}     & \multicolumn{1}{c}{\textbf{100.00}}    & \multicolumn{1}{c}{\textbf{100.00}}   & 65.00                & \multicolumn{1}{c}{76.67}     & \multicolumn{1}{c}{66.67}     & \underline{93.33} &\underline{91.67} &\underline{85.00}\\
Complete Information             & \textbf{100.00}                & \multicolumn{1}{c}{\textbf{96.67}}  & \multicolumn{1}{c}{\textbf{93.33}} & 80.00                & \multicolumn{1}{c}{68.33}     & \multicolumn{1}{c}{70.00}  & \underline{93.33} &\underline{91.67} &\underline{85.00}\\
Within Current City              & \textbf{100.00}                  & \multicolumn{1}{c}{\textbf{100.00}}    & \multicolumn{1}{c}{\textbf{100.00}}   & 88.33                & \underline{91.67}     & \underline{93.33} & \underline{93.33} &\underline{91.67} &85.00\\
Reasonable City Route            & \textbf{100.00}                  & \multicolumn{1}{c}{\textbf{100.00}}    & \multicolumn{1}{c}{\textbf{100.00}}   & \underline{96.67}                & \multicolumn{1}{c}{86.67}  & \underline{95.00}  & 93.33 &\underline{91.67} &85.00\\
Diverse Restaurants              & \textbf{100.00}                  & \multicolumn{1}{c}{\textbf{100.00}}    & \multicolumn{1}{c}{\textbf{100.00}}   & 83.33                   & \multicolumn{1}{c}{88.33}     & \underline{91.67}     & \underline{93.33} &\underline{91.67} &85.00 \\
Diverse Attractions              & \textbf{100.00}                  & \multicolumn{1}{c}{\textbf{100.00}}    & \multicolumn{1}{c}{\textbf{100.00}}   &86.67                & \underline{98.33}  & \underline{98.33}    & \underline{93.33} &91.67 &85.00\\
Non-conf. Transportation         & \textbf{100.00}                & \multicolumn{1}{c}{\textbf{98.33}}  & \underline{95.00}    & 81.67                   & \underline{96.67}  & \textbf{96.67} & \underline{93.33} &91.67&85.00\\
Minimum Nights Stay              & \textbf{98.33}                & \multicolumn{1}{c}{\textbf{93.33}}  & \multicolumn{1}{c}{\underline{86.67}} & 90.00                   & 86.67     & \textbf{90.00} & \underline{93.33} &\underline{91.67}&85.00  \\ \hline
\multicolumn{10}{c}{Hard Constraint}  \\\hline
Budget                           & \textbf{100.00}                   & \multicolumn{1}{c}{\textbf{93.33}}     & \multicolumn{1}{c}{\textbf{88.33}}    & 46.67                    & \multicolumn{1}{c}{38.33}   & \multicolumn{1}{c}{43.33}    &\underline{85.00} &\underline{90.00}&\underline{80.00}\\
Room Rule                        & -                    & \multicolumn{1}{c}{\textbf{90.00}}     & \multicolumn{1}{c}{\textbf{87.72}} & -                    & \multicolumn{1}{c}{\underline{50.00}}      & \multicolumn{1}{c}{59.65}   &- &\textbf{90.00} &\underline{84.21}\\
Cuisine                          & -                    & \underline{81.82}  & \multicolumn{1}{c}{\textbf{84.62}} & -                    & \multicolumn{1}{c}{45.45}   & \multicolumn{1}{c}{\underline{61.54} }    &- &\textbf{100.00} &\textbf{84.62}\\
Room Type                        & -                    & \multicolumn{1}{c}{\textbf{83.33}}  & \multicolumn{1}{c}{\textbf{86.96}} & -                    & \multicolumn{1}{c}{\underline{38.89}}   & \multicolumn{1}{c}{54.35}   &- &\textbf{83.33} &\underline{84.78}\\
Transportation                   & -                    & \multicolumn{1}{c}{-}      & \multicolumn{1}{c}{\textbf{90.20}} & -                    & \multicolumn{1}{c}{-}      & \multicolumn{1}{c}{58.82}   &- & - &\underline{86.27}\\\hline
\multicolumn{10}{c}{Final Pass Rate}  \\\hline
Final Pass Rate                  & \textbf{98.33}                & \multicolumn{1}{c}{\textbf{93.33}}  & \multicolumn{1}{c}{\textbf{83.33}} & 43.33                 & \multicolumn{1}{c}{20.00}   & \multicolumn{1}{c}{26.67}     &\underline{85.00}  &\underline{90.00} &\underline{80.00} \\\hline
\end{tabular}
}
\label{constrainttype}
\end{table*}

Table \ref{constrainttype} shows a detailed performance comparison of the four methods under different commonsense and hard constraints on three difficulty levels. With respect to the commonsense constraints, our method achieves near-perfect satisfaction rates on easy, medium, and hard levels. Specifically, it reaches a perfect 100\% on constraints such as \textit{Within Sandbox}, \textit{Within Current City}, \textit{Reasonable City Route}, \textit{Diverse Restaurants}, and \textit{Diverse Attractions} at all difficulty levels, demonstrating robust reasoning over context-dependent and routine constraints. In comparison, llm-rwplanning generally ranks second, with slightly lower commonsense pass rates than Behavior Forest. Notably, its commonsense constraint performance remains largely uniform across different constraint types. This is because llm-rwplanning adopts a hard feasibility enforcement strategy, in which plans are generated only if all commonsense constraints are satisfied. When any commonsense constraint cannot be met, the method fails to produce a feasible plan. As a result, only fully compliant plans are counted, leading to identical satisfaction rates across individual commonsense constraints. While this design enforces strict feasibility, it limits flexibility in handling partially satisfied constraints and reduces robustness in more complex or interdependent settings. PMC exhibits greater variability across different constraints: under the hard level, the constraint \textit{Within Sandbox} is satisfied at 66.67\%, whereas the constraint \textit{Diverse Attractions} achieves 98.33\%.

For hard constraints, Behavior Forest also demonstrates substantial advantages. It achieves the best performance on most hard constraints, such as \textit{Budget} (100\%, 93.33\%, 88.33\% for easy, medium, and hard, respectively), and \textit{Room Rule} (90\%, 87.72\% for medium and hard, respectively). Llm-rwplanning shows competitive performance on certain hard constraints, e.g., \textit{Cuisine} reaches 100\% and \textit{Room Type} reaches 83.33\% at the medium difficulty level, but generally falls behind the behavior forest at the hard difficulty level. The PMC scores remain low across most hard constraints, highlighting its limited ability to handle multi-step, interdependent planning under strict requirements. The final pass rates further corroborate this trend: Behavior Forest reaches 98.33\%, 93.33\%, and 83.33\% for easy, medium, and hard levels, significantly outperforming both PMC and llm-rwplanning.

\textit{\textbf{Summary}} Behavior Forest achieves the highest pass rates across most commonsense and hard constraints at all difficulty levels. Its final pass rates are 98.33\%, 93.33\%, and 83.33\% for easy, medium, and hard tasks, respectively, while llm-rwplanning generally ranks second and PMC performs substantially lower.

\subsection{Performance Analysis Across Different Difficulty Levels}

\begin{table*}[]
\centering
\caption{Performance comparison of different planning methods across varying difficulty levels. The best results within each difficulty level are highlighted in \textbf{bold}, while the second-best scores are underlined using \underline{underline}.}
\scalebox{1.2}{
\begin{tabular}{cccccccc} \hline
\multicolumn{2}{c}{\multirow{2}{*}{Methods}} & \multicolumn{1}{c}{\multirow{2}{*}{Delivery Rate}} & \multicolumn{2}{c}{Commonsense Pass Rate} & \multicolumn{2}{c}{Hard Pass Rate}    & \multicolumn{1}{c}{\multirow{2}{*}{Final Pass Rate}}  \\ \cmidrule(lr){4-5}\cmidrule(lr){6-7}
\multicolumn{2}{c}{}  & \multicolumn{1}{c}{}                               & \multicolumn{1}{c}{Micro} & \multicolumn{1}{c}{Macro} & \multicolumn{1}{c}{Micro} & \multicolumn{1}{c}{Macro} & \multicolumn{1}{c}{}  \\ \hline
\multicolumn{1}{c}{\multirow{4}{*}{Easy}}  & \multicolumn{1}{c}{ReAct\cite{ReAct}}        & \underline{96.67}                          & \multicolumn{1}{c}{60.42}    & \multicolumn{1}{c}{5.00}     & \multicolumn{1}{c}{5.00}     & \multicolumn{1}{c}{5.00}     & \multicolumn{1}{c}{1.67} \\
\multicolumn{1}{c}{}                        & \multicolumn{1}{c}{EvoAgent\cite{yuan2025evoagent}}     & \textbf{100.00}                           & \underline{97.08} & \multicolumn{1}{c}{18.53} & \multicolumn{1}{c}{66.67} & \multicolumn{1}{c}{66.67} & \multicolumn{1}{c}{6.67} \\
\multicolumn{1}{c}{}                         & \multicolumn{1}{c}{PMC\cite{Multi-ConstraintsCollaborative}}          & \textbf{100.00}                            & \multicolumn{1}{c}{83.96} & \multicolumn{1}{c}{65.00}    & \multicolumn{1}{c}{46.67}    & \multicolumn{1}{c}{48.33}    & \multicolumn{1}{c}{43.33}  \\
  \multicolumn{1}{c}{}  & \multicolumn{1}{c}{llm-rwplanning\cite{RealWorld}}          & \multicolumn{1}{c}{93.33}                             & \multicolumn{1}{c}{93.33} & \underline{93.33}    & \underline{85.00} & \underline{85.00}    & \underline{85.00} \\
 & \multicolumn{1}{c}{\textbf{Behavior Forest}} & \multicolumn{1}{c}{\textbf{100.00}} & \multicolumn{1}{c}{\textbf{99.79}} & \multicolumn{1}{c}{\textbf{98.33}} & \multicolumn{1}{c}{\textbf{100.00}} & \multicolumn{1}{c}{\textbf{98.33}} & \multicolumn{1}{c}{\textbf{98.33}} \\ \hline
\multicolumn{1}{c}{\multirow{4}{*}{Medium}}  & \multicolumn{1}{c}{ReAct\cite{ReAct}}        & \multicolumn{1}{c}{91.67}                          & \multicolumn{1}{c}{50.42} & \multicolumn{1}{c}{3.33}  & \multicolumn{1}{c}{2.50}  & \multicolumn{1}{c}{3.33}  & \multicolumn{1}{c}{1.67} \\
\multicolumn{1}{c}{}                        & \multicolumn{1}{c}{EvoAgent\cite{yuan2025evoagent}}     & \textbf{100.00}                            & \multicolumn{1}{c}{76.25}  & \multicolumn{1}{c}{16.67} & \multicolumn{1}{c}{53.33} & \multicolumn{1}{c}{15.26} & \multicolumn{1}{c}{3.33}  \\ 
\multicolumn{1}{c}{}    & \multicolumn{1}{c}{PMC\cite{Multi-ConstraintsCollaborative}}          & \underline{96.00}                             & \multicolumn{1}{c}{86.67}  & \multicolumn{1}{c}{25.00}    & \multicolumn{1}{c}{41.67}    & \multicolumn{1}{c}{61.67}    & \multicolumn{1}{c}{20.00}  \\
  \multicolumn{1}{c}{}  & \multicolumn{1}{c}{llm-rwplanning\cite{RealWorld}}          & \multicolumn{1}{c}{91.67}                             & \underline{91.67} & \underline{91.67}    & \multicolumn{1}{c}{\textbf{90.83}} & \underline{90.00}    & \underline{90.00} \\
& \multicolumn{1}{c}{\textbf{Behavior Forest}} & \multicolumn{1}{c}{\textbf{100.00}} & \multicolumn{1}{c}{\textbf{98.54}} & \multicolumn{1}{c}{\textbf{95.00}} & \underline{89.17} & \multicolumn{1}{c}{\textbf{93.33}} & \multicolumn{1}{c}{\textbf{93.33}}  \\ \hline
\multicolumn{1}{c}{\multirow{4}{*}{Hard}}  & \multicolumn{1}{c}{ReAct\cite{ReAct}}        & \multicolumn{1}{c}{66.67}                          & \multicolumn{1}{c}{49.38} & \multicolumn{1}{c}{0.00}     & \multicolumn{1}{c}{1.25}  & \multicolumn{1}{c}{0.00}     & \multicolumn{1}{c}{0.00}  \\
\multicolumn{1}{c}{}                        & \multicolumn{1}{c}{EvoAgent\cite{yuan2025evoagent}}     & \textbf{100.00}                            & \multicolumn{1}{c}{55.42} & \multicolumn{1}{c}{10.10}  & \multicolumn{1}{c}{24.17} & \multicolumn{1}{c}{3.33}  & \multicolumn{1}{c}{3.33}  \\ 
 \multicolumn{1}{c}{}  & \multicolumn{1}{c}{PMC\cite{Multi-ConstraintsCollaborative}}          &\underline{96.00}                             & \multicolumn{1}{c}{\underline{87.71}} & \multicolumn{1}{c}{58.33}    & \multicolumn{1}{c}{54.58} & \multicolumn{1}{c}{40.00}    & \multicolumn{1}{c}{26.67} \\
  \multicolumn{1}{c}{}  & \multicolumn{1}{c}{llm-rwplanning\cite{RealWorld}}          & \multicolumn{1}{c}{85.00}                             & 85.00 & \underline{85.00}    & \underline{83.75} & \underline{80.00}    & \underline{80.00} \\
& \multicolumn{1}{c}{\textbf{Behavior Forest}} & \multicolumn{1}{c}{\textbf{100.00}} & \multicolumn{1}{c}{\textbf{96.88}} & \multicolumn{1}{c}{\textbf{90.00}} & \multicolumn{1}{c}{\textbf{87.92}} & \multicolumn{1}{c}{\textbf{83.33}} & \multicolumn{1}{c}{\textbf{83.33}} \\ \hline
\end{tabular}
}
\label{difficult}
\end{table*}

Based on the experimental results presented in Table \ref{difficult}, Behavior Forest consistently achieves the highest final pass rates across all difficulty levels, outperforming all baseline methods. At the easy level, Behavior Forest achieves a final pass rate of 98.33\%, surpassing llm-rwplanning (85\%), PMC (43.33\%), EvoAgent (6.67\%), and ReAct (1.67\%). Behavior Forest also maintains the highest delivery rate (100\%) and near-perfect commonsense pass rates as well as perfect hard pass rates, demonstrating its effectiveness in satisfying both commonsense and hard constraints. 

At the medium level, Behavior Forest maintains clear advantages, with a final pass rate of 93.33\%, outperforming llm-rwplanning (90\%), PMC (20\%), EvoAgent (3.33\%), and ReAct (1.67\%). Although the performance of most methods declines compared with easy tasks, Behavior Forest maintains high commonsense pass rates and hard constraint satisfaction. The difference in performance between our method and the baselines increases, highlighting the ability of the framework to handle interdependent constraints as the task difficulty increases.

At the hard level, Behavior Forest still leads, achieving a final pass rate of 83.33\%, while ReAct falls to 0\%. Although the performance declines for all methods under maximum difficulty, Behavior Forest maintains superior satisfaction across both commonsense and hard constraints, demonstrating that parallel decoupling and hierarchical planning effectively reduce cascading errors and constraint interference even in the most challenging multi-constraint settings.

\textit{\textbf{Summary}} Behavior Forest achieves the highest final pass rates across all difficulty levels (98.33\% for easy, 93.33\% for medium, and 83.33\% for hard tasks), outperforming other baselines, while maintaining a 100\% delivery rate.

\subsection{Analysis of Error Types}

\begin{figure}
    \centering
    \includegraphics[width=1\linewidth]{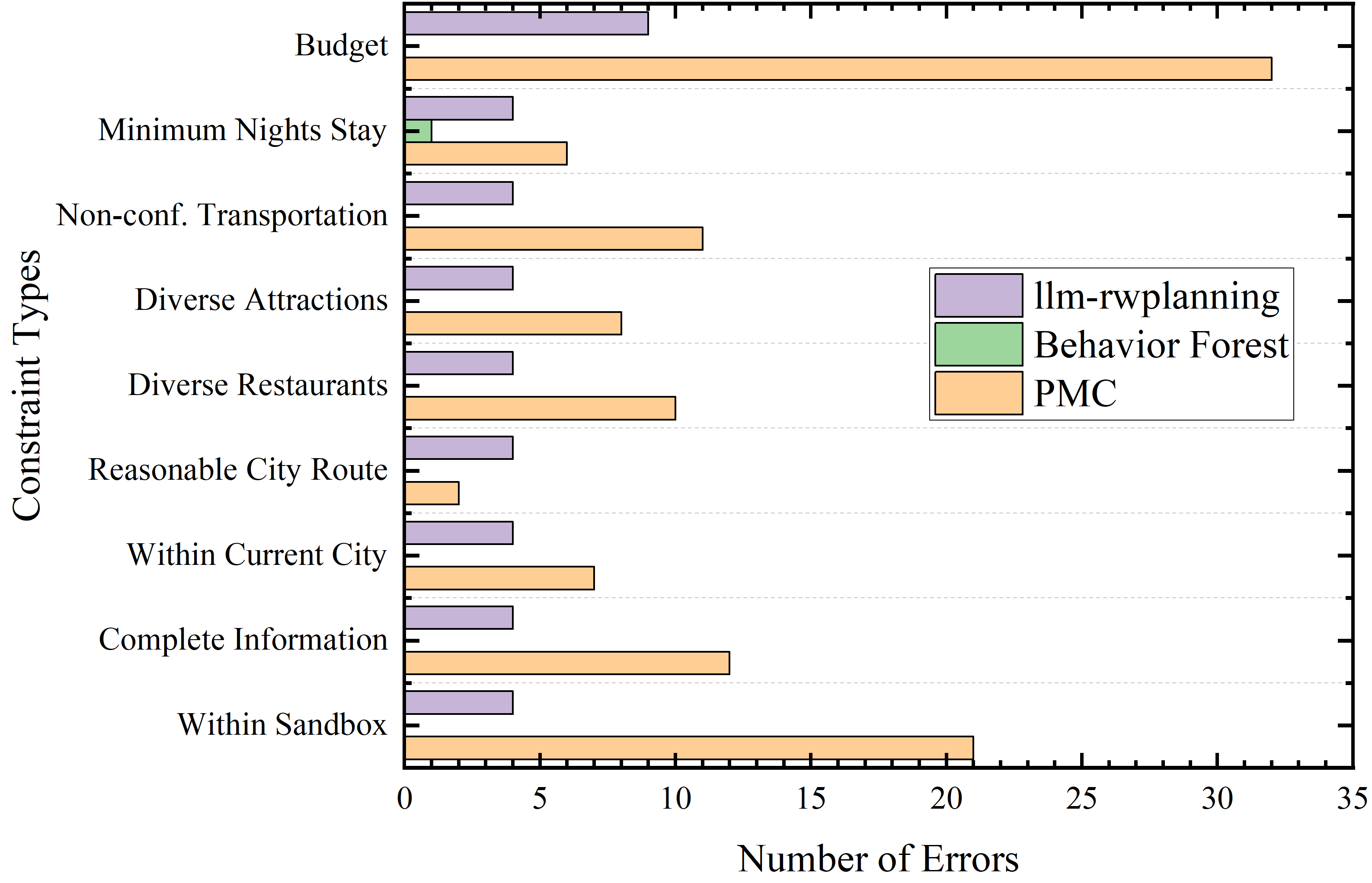}
    \caption{Violations for different constraint types across three methods (Behavior Forest, llm-rwplanning, and PMC) at the easy difficulty level.}
    \label{easy}
\end{figure}

\begin{figure}
    \centering
    \includegraphics[width=1\linewidth]{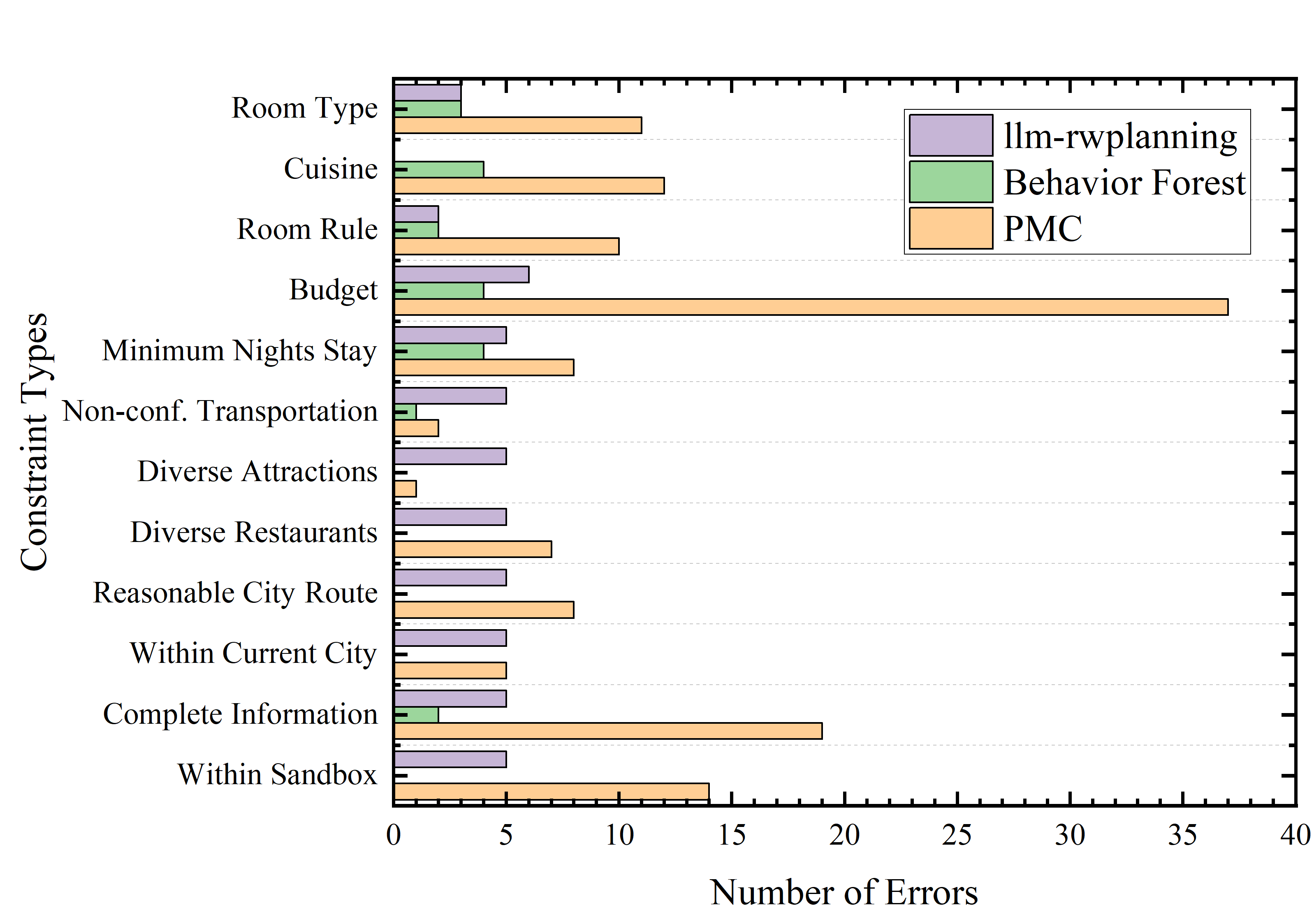}
    \caption{Violations for different constraint types across three methods (Behavior Forest, llm-rwplanning, and PMC) at the medium difficulty level.}
    \label{Medium}
\end{figure}

\begin{figure}
    \centering
    \includegraphics[width=1\linewidth]{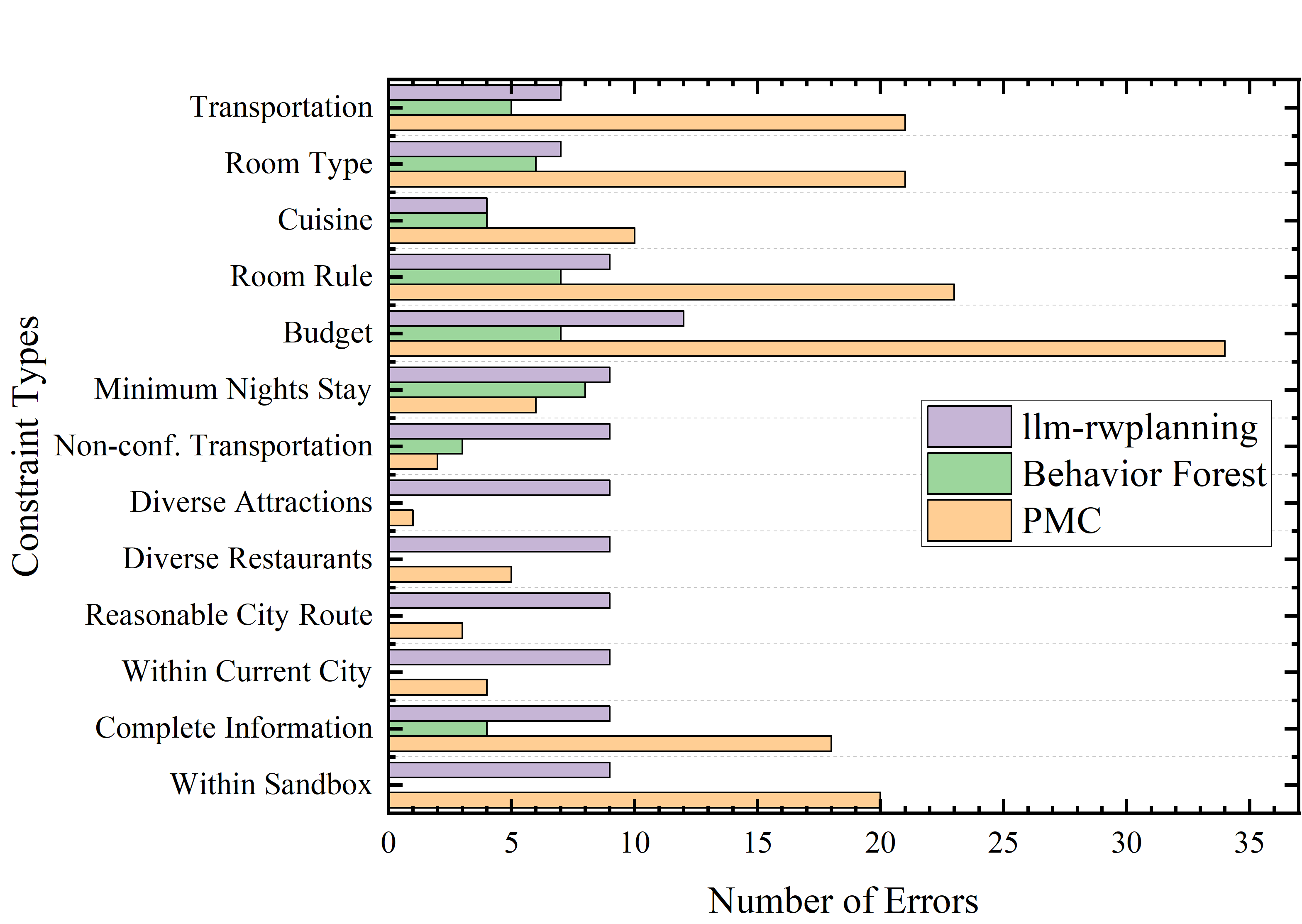}
    \caption{Violations for different constraint types across three methods (Behavior Forest, llm-rwplanning, and PMC) at the hard difficulty level.}
    \label{Hard}
\end{figure}

The numbers of constraint violations for Behavior Forest, llm-rwplanning, and PMC at the easy task level are shown in Figure \ref{easy}. PMC has the most violations in most constraint types, such as \textit{Budget}, and \textit{Within Sandbox}. The llm-rwplanning performs better than PMC but worse than Behavior Forest in terms of the constraints. Behavior Forest consistently shows the lowest violation across all constraint types, demonstrating its better constraint adherence.

At the medium task level (Figure \ref{Medium}), the performance gap among the methods becomes more pronounced. PMC shows a substantial number of violations, especially for the \textit{Budget} constraint. The increase in these violations as constraint interdependencies grow suggests that PMC’s performance deteriorates under more complex conditions. The llm-rwplanning experiences a moderate rise in violations but remains clearly more controlled than PMC. Behavior Forest demonstrates the strongest robustness, with only small increments across a few constraints despite the increased task complexity.

At the hard task level (Figure \ref{Hard}), PMC exhibits extensive violations across some constraint types, such as \textit{Budget}, \textit{Room Rule}, and \textit{Room Type}. The llm-rwplanning exhibits modest violation growth, remaining clearly less affected than PMC.
Behavior Forest demonstrates strong stability, allowing only small increases and maintaining the lowest levels despite the increased task difficulty.

\textit{\textbf{Summary}} Behavior Forest exhibits the fewest constraint violations across all difficulty levels, demonstrating stronger constraint adherence and robustness than the other baselines.

\subsection{Ablation Study}

\begin{table*}[h]
\centering
\caption{Ablation study of Behavior Forest on the TravelPlanner benchmark, evaluating the contributions of the coordination mechanism, candidate solution reranking (Rerank), and heuristic scoring. The best scores in each column are highlighted in \textbf{bold}, while the second-best scores are underlined using \underline{underline}.}
\scalebox{1.2}{
\begin{tabular}{ccccccc}\hline
\multirow{2}{*}{Method} & \multirow{2}{*}{Delivery Rate} & \multicolumn{2}{c}{Commonsense Pass Rate} & \multicolumn{2}{c}{Hard Pass Rate} & \multirow{2}{*}{Final Pass Rate}\\ 
\cmidrule(lr){3-4}\cmidrule(lr){5-6}
 &  & Micro & Macro & Micro & Macro & \\\hline
w/o\ coordination mechanism     & \textbf{100.00} & 83.96 & 76.67 & 73.57 & 75.56 & 75.56\\
w/o\ Rerank          & \textbf{100.00} & 92.43 & 85.56 & 82.62 & 77.78 & 77.78\\
w/o\ Heuristic scoring & \textbf{100.00} & \underline{95.28} &\underline{87.22} & \underline{86.19} & \underline{80.56} & \underline{80.56}\\
\textbf{Behavior Forest}  & \textbf{100.00} & \textbf{98.40} & \textbf{94.44} & \textbf{90.00} & \textbf{91.67} & \textbf{91.67}\\ \hline
\end{tabular}
}
\label{ablation}
\end{table*}

The ablation study in Table \ref{ablation} demonstrates the contribution of each core component to Behavior Forest. Across all the metrics, the full framework consistently outperforms its ablated variants. Removing the coordination mechanism leads to the largest performance drop, with the final pass rate decreasing from 91.67\% to 75.56\%. In addition, the commonsense macro pass rate and hard macro pass rate decrease from 94.44\% to 76.67\% and from 91.67\% to 75.56\%, respectively. This significant decrease is primarily due to the removal of the coordination mechanism, which ensures that the plans generated by all behavior trees satisfy the global constraints. Without this mechanism, it becomes difficult to guarantee global consistency, causing multiple violations and reducing overall performance.

Excluding the candidate solution reranking mechanism results in moderate reductions, with the final pass rate falling to 77.78\%, and corresponding decreases in micro/macro metrics for both commonsense and hard constraints. Similarly, removing heuristic scoring reduces the final pass rate to 80.56\%. These declines can be attributed to the role of both components in guiding the selection within the candidate solution pool. Without them, the framework randomly selects from the candidate solutions, making it difficult to choose solutions that satisfy current requirements, thereby lowering overall planning performance.

\textit{\textbf{Summary}} The ablation study demonstrates that each component—coordination mechanism, candidate solution reranking, and heuristic scoring—significantly contributes to overall performance. Their removal leads to a decline across all metrics, confirming the synergistic effect of all three components.

\subsection{Performance Comparison with Different LLMs}

\begin{table*}[]
\centering
\caption{Performance comparison of Behavior Forest and baseline methods across different LLMs (Mistral-7B, GPT-3.5, and DeepSeek-V3) on the TravelPlanner benchmark. \textbf{Bold} values denote the best performance in each column for the corresponding LLMs, while the second-best scores are underlined using \underline{underline}.}
\scalebox{1.1}{
\begin{tabular}{cccccccc}\hline
\multicolumn{1}{c}{\multirow{2}{*}{Model}}       & \multicolumn{1}{c}{\multirow{2}{*}{Method}} & \multicolumn{1}{c}{\multirow{2}{*}{Delivery Rate}} & \multicolumn{2}{c}{Commonsense Pass Rate}             & \multicolumn{2}{c}{Hard Pass Rate}                    & \multicolumn{1}{c}{\multirow{2}{*}{Final Pass Rate}} \\ \cmidrule(lr){4-5}\cmidrule(lr){6-7}
\multicolumn{1}{c}{}                             & \multicolumn{1}{c}{}                        & \multicolumn{1}{c}{}                               & \multicolumn{1}{c}{Micro} & \multicolumn{1}{c}{Macro} & \multicolumn{1}{c}{Micro} & \multicolumn{1}{c}{Macro} & \multicolumn{1}{c}{}           \\ \hline
\multicolumn{1}{c}{\multirow{5}{*}{Mistral-7B}}  & \multicolumn{1}{c}{Direct\cite{TravelPlanner}}                  & \textbf{100.00}                            & 55.56  & \multicolumn{1}{c}{1.67}   & \multicolumn{1}{c}{1.19}   & \multicolumn{1}{c}{0.00}   & \multicolumn{1}{c}{0.00}               \\
\multicolumn{1}{c}{}                             & \multicolumn{1}{c}{CoT\cite{wei2022chain}}                     & \textbf{100.00}                            & \multicolumn{1}{c}{53.47}  & \multicolumn{1}{c}{0.00}     & \multicolumn{1}{c}{2.38}   & \multicolumn{1}{c}{0.00}   & \multicolumn{1}{c}{0.00}               \\
\multicolumn{1}{c}{}                             & \multicolumn{1}{c}{PMC\cite{Multi-ConstraintsCollaborative}}                & \underline{77.78}                            & 42.08  & 21.67   & 24.29   & \multicolumn{1}{c}{19.44}     & \multicolumn{1}{c}{13.89}               \\
\multicolumn{1}{c}{}                             & \multicolumn{1}{c}{llm-rwplanning\cite{RealWorld}}                   & 56.67                           & \multicolumn{1}{c}{\underline{56.67}}  & \multicolumn{1}{c}{\underline{56.67}}   & \multicolumn{1}{c}{\underline{33.81}}   & \multicolumn{1}{c}{\underline{33.33}}   & \multicolumn{1}{c}{\underline{33.33}}                \\
\multicolumn{1}{c}{}                             & \multicolumn{1}{c}{\textbf{Behavior Forest}} & \multicolumn{1}{c}{\textbf{100.00}} & \multicolumn{1}{c}{\textbf{62.92}} & \multicolumn{1}{c}{\textbf{58.89}} & \multicolumn{1}{c}{\textbf{34.76}} & \multicolumn{1}{c}{\textbf{34.44}} & \multicolumn{1}{c}{\textbf{34.44}}             \\ \hline
\multicolumn{1}{c}{\multirow{5}{*}{GPT-3.5}}     & \multicolumn{1}{c}{Direct\cite{TravelPlanner}}                  & \textbf{100.00}                            & \multicolumn{1}{c}{48.61}  & \multicolumn{1}{c}{2.78}   & \multicolumn{1}{c}{8.81}   & \multicolumn{1}{c}{2.78}   & \multicolumn{1}{c}{0.00}                \\
\multicolumn{1}{c}{}                             & \multicolumn{1}{c}{CoT\cite{wei2022chain}}                     & \textbf{100.00}                            & \multicolumn{1}{c}{50.21}  & \multicolumn{1}{c}{1.67}   & \multicolumn{1}{c}{7.62}   & \multicolumn{1}{c}{1.67}   & \multicolumn{1}{c}{0.00}                  \\
\multicolumn{1}{c}{}                             & \multicolumn{1}{c}{PMC\cite{Multi-ConstraintsCollaborative}}                & \underline{91.67}                            & \underline{78.89}  & 41.67   & 39.76  & 40.00   & 28.89                   \\
\multicolumn{1}{c}{}                             & \multicolumn{1}{c}{llm-rwplanning\cite{RealWorld}}                   & 77.78                           & \multicolumn{1}{c}{77.78}  & \multicolumn{1}{c}{\underline{77.78}}   & \multicolumn{1}{c}{\underline{76.19}}   & \multicolumn{1}{c}{\underline{75.56}}   & \multicolumn{1}{c}{\underline{75.56}}                 \\
\multicolumn{1}{c}{}                             & \multicolumn{1}{c}{\textbf{Behavior Forest}} & \multicolumn{1}{c}{\textbf{100.00}} & \multicolumn{1}{c}{\textbf{89.31}} & \multicolumn{1}{c}{\textbf{85.56}} & \multicolumn{1}{c}{\textbf{79.29}} & \multicolumn{1}{c}{\textbf{80.56}} & \multicolumn{1}{c}{\textbf{76.11}}             \\ \hline
\multicolumn{1}{c}{\multirow{5}{*}{DeepSeek-V3}} & \multicolumn{1}{c}{Direct\cite{TravelPlanner}}                  & \textbf{100.00}                            & \multicolumn{1}{c}{72.78}  & \multicolumn{1}{c}{14.44}  & \multicolumn{1}{c}{15.95}  & \multicolumn{1}{c}{8.89}   & \multicolumn{1}{c}{1.67}               \\
\multicolumn{1}{c}{}                             & \multicolumn{1}{c}{CoT\cite{wei2022chain}}                     & \textbf{100.00}                            & \multicolumn{1}{c}{68.40}  & \multicolumn{1}{c}{8.33}   & \multicolumn{1}{c}{13.57}  & \multicolumn{1}{c}{9.44}   & \multicolumn{1}{c}{1.11}               \\
\multicolumn{1}{c}{}                             & \multicolumn{1}{c}{PMC\cite{Multi-ConstraintsCollaborative}}                & \underline{97.33}                            & 86.11 & 49.44 & 49.76 & 50.00 & 30.00   \\
\multicolumn{1}{c}{}                             & \multicolumn{1}{c}{llm-rwplanning\cite{RealWorld}}                   & 90.00                          & \multicolumn{1}{c}{\underline{90.00}} & \multicolumn{1}{c}{\underline{90.00}}  & \multicolumn{1}{c}{\underline{85.95}}  & \multicolumn{1}{c}{\underline{85.00}}  & \multicolumn{1}{c}{\underline{85.00}}            \\
\multicolumn{1}{c}{}                             & \multicolumn{1}{c}{\textbf{Behavior Forest}} & \multicolumn{1}{c}{\textbf{100.00}} & \multicolumn{1}{c}{\textbf{98.40}} & \multicolumn{1}{c}{\textbf{94.44}} & \multicolumn{1}{c}{\textbf{90.00}} & \multicolumn{1}{c}{\textbf{91.67}} & \multicolumn{1}{c}{\textbf{91.67}}  \\\hline
\end{tabular}
}
\label{diffLLMs}
\end{table*}

The results in Table \ref{diffLLMs} demonstrate that Behavior Forest consistently outperforms all baseline methods for different LLMs. With respect to the DeepSeek-V3 model, Behavior Forest achieves a final pass rate of 91.67\%, substantially higher than that of other methods. Similarly, with GPT-3.5, Behavior Forest reaches 76.11\%, while llm-rwplanning reaches 75.56\% and PMC reaches 28.89\%.  Even with the smaller Mistral-7B model, Behavior Forest achieves a final pass rate of 34.44\%, while Direct and CoT methods fail to produce plans, highlighting the robustness of our method. 

Behavior Forest maintains a perfect delivery rate (100\%) in all LLMs, demonstrating reliable plan generation regardless of model size. Compared with the strongest baseline, llm-rwplanning, Behavior Forest achieves higher pass rates on both commonsense and hard constraints, indicating its superior capability in complex multi-constraint reasoning. 
Moreover, its performance advantage is evident not only on micro-level metrics but also on macro-level evaluation, confirming that Behavior Forest effectively balances different constraint types and generalizes well across LLM configurations.

\textit{\textbf{Summary}} Behavior Forest achieves higher final pass rates across all LLMs (91.67\% with DeepSeek-V3, 76.11\% with GPT-3.5, and 34.44\% with Mistral-7B) than baseline methods, and also outperforms the baselines on other key metrics, demonstrating the method's robustness and effectiveness.

\subsection{Case Studies}

\begin{figure*}
    \centering
    \includegraphics[width=1\linewidth]{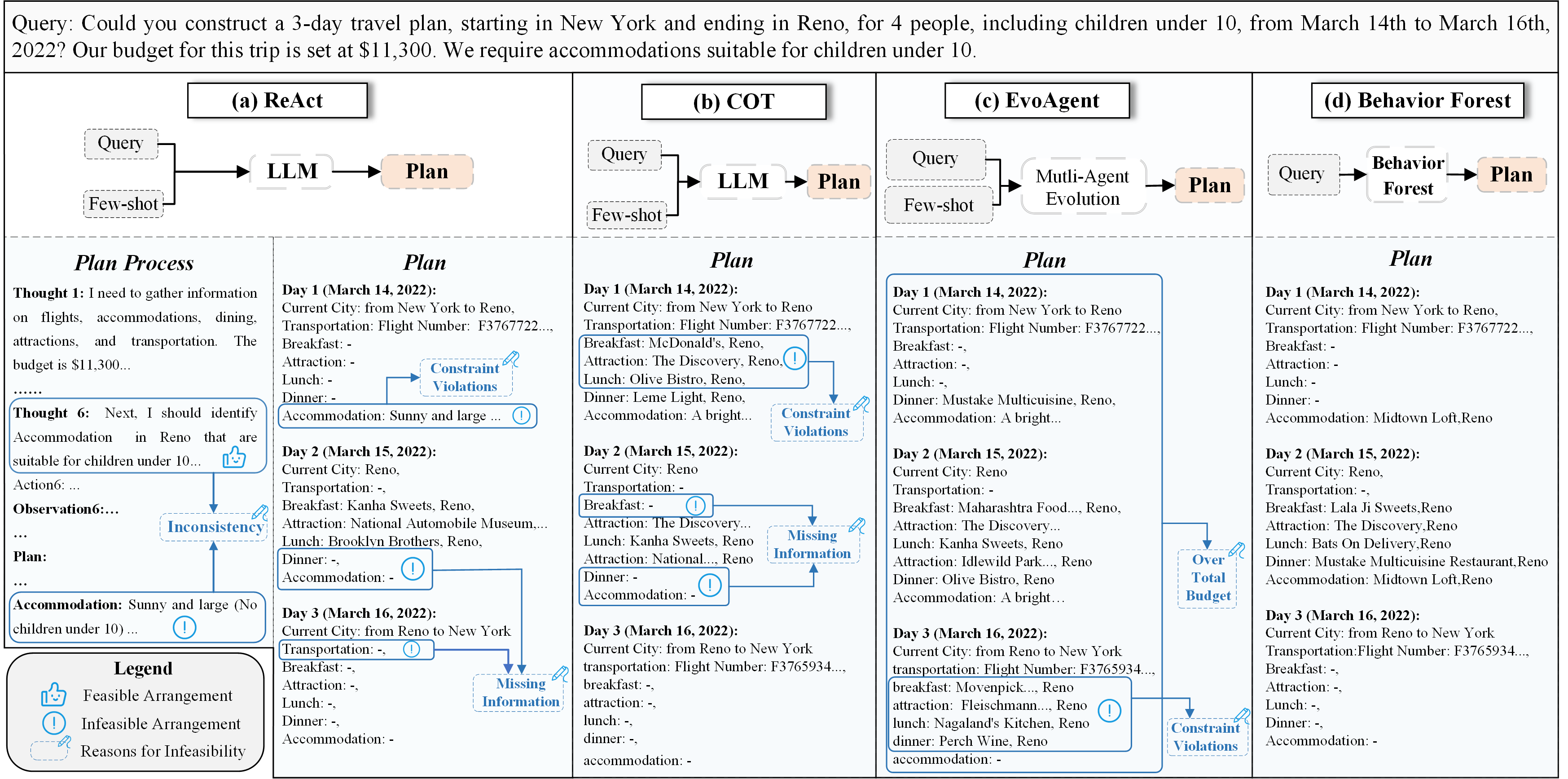}
    \caption{Case study comparing Behavior Forest with three representative baselines: ReAct, CoT, and EvoAgent. ReAct suffers from inconsistency between reasoning and execution, whereas CoT’s exemplar dependence leads to planning errors and omissions. EvoAgent alleviates incompleteness but introduces constraint violations due to excessive diversity exploration. In contrast, Behavior Forest achieves valid plans by decoupling multi-constraint reasoning into structured subspaces, demonstrating superior reliability in travel planning.}
    \label{case}
\end{figure*}

To further demonstrate the advantages of Behavior Forest, we conduct a case study, as illustrated in Figure~\ref{case}. Among them, ReAct represents parameterized-knowledge methods, whereas CoT and EvoAgent are prior-augmented approaches.

As shown in Figure~\ref{case}(a), ReAct generates a travel plan based on internal knowledge of LLM, but reveals a clear inconsistency between reasoning and execution. Although the LLM correctly recognizes the need for suitable accommodation in Reno for children under 10, the final plan contradicts this reasoning by selecting Sunny and large (No children under 10). Furthermore, critical elements such as dinner and accommodation on Day 2 are missing, indicating unstable alignment between internal reasoning and external planning.

The plan produced by CoT is shown in Figure~\ref{case}(b). Although it benefits from few-shot reasoning, its excessive reliance on exemplars limits adaptability. For example, flight F3767722 arrives in Reno the following day, rendering the breakfast and attraction arrangements invalid. In addition, omissions in Day 2’s meal planning suggest that CoT inherits contextual biases from demonstrations, ultimately causing planning failures.

As shown in Figure~\ref{case}(c), EvoAgent employs multi-agent evolution to generate plans, mitigating the incompleteness issues found in previous methods. However, excessive pursuit of diversity causes constraint violations. The generated plan exceeds the budget. It schedules a flight on Day 3, which conflicts with subsequent restaurant and attraction arrangements. The results show that EvoAgent struggles to simultaneously achieve diverse exploration and satisfy constraints.

As shown in Figure~\ref{case}(d),  Behavior Forest generates a valid plan that meets user requirements, including budget and accommodation suitability. By leveraging the forest structure, Behavior Forest decouples complex multi-constraint reasoning into structured and manageable subspaces, enabling each subtask to operate within its respective constraint domain. This structured decoupling effectively reduces the reasoning burden of LLMs and ensures inter-constraint consistency, leading to a more reliable planning process.

\textit{\textbf{Summary}} The case study demonstrates that Behavior Forest effectively alleviates the reasoning–execution inconsistencies and constraint violations observed in existing methods. By leveraging the forest structure, it decouples multi-constraint planning into structured and independent subtasks, allowing each to operate within its specific constraint domain. This decoupling reduces the reasoning burden of LLMs and enhances the validity of planning.

\subsection{Analysis of Runtime}

\begin{table}[]
\centering
\caption{Comparison of runtime, token consumption, and final pass rate (\%) for Behavior Forest and baseline methods on the TravelPlanner benchmark. BF-Seq is our method without parallel planning, in which subtasks are executed sequentially. \textbf{Bold} indicates the best performance in each column, while the second-best scores are underlined using \underline{underline}.}
\scalebox{0.95}{
\begin{tabular}{cccc} \hline
\multicolumn{1}{c}{Method}   & \multicolumn{1}{c}{Runtime(s)} & \multicolumn{1}{c}{Token consumption} & \multicolumn{1}{c}{Final Pass Rate} \\\hline
\multicolumn{1}{c}{Direct\cite{TravelPlanner}}          & \underline{231.46}     &  \textbf{768 / 635}       & \multicolumn{1}{c}{1.67} \\
\multicolumn{1}{c}{CoT\cite{wei2022chain}}             & \multicolumn{1}{c}{305.71}     &  4562 / \underline{1127}       & \multicolumn{1}{c}{1.11}   \\
\multicolumn{1}{c}{PMC\cite{Multi-ConstraintsCollaborative}}           & 451.02     &   5632 / 2856       & \multicolumn{1}{c}{30.00}  \\
\multicolumn{1}{c}{llm-rwplanning\cite{RealWorld}}        & 395.73      &   14039 / 6383      & \underline{85.00}  \\
\multicolumn{1}{c}{BF-Seq}     & \multicolumn{1}{c}{428.25}      & -        & -  \\
\multicolumn{1}{c}{\textbf{Behavior Forest}} & \multicolumn{1}{c}{\textbf{122.13}}   & \underline{3210} / 1586           & \multicolumn{1}{c}{\textbf{91.67}}  \\\hline
\end{tabular}
}
\label{runtime}
\end{table}

The results in Table \ref{runtime} demonstrate the clear advantages of Behavior Forest in balancing runtime, token consumption, and final pass rate. Among all the methods, Behavior Forest achieves the shortest runtime of 122.13 s, making it the fastest approach overall, while simultaneously obtaining the highest final pass rate of 91.67\%. In terms of token consumption, Behavior Forest uses 3210 input tokens and 1586 output tokens, which is substantially lower than llm-rwplanning and PMC but achieves significantly better performance. Although Direct and CoT consume fewer tokens, their final pass rates remain extremely low (1.67\% and 1.11\%, respectively). Compared with llm-rwplanning, Behavior Forest not only improves the final pass rate from 85.00\% to 91.67\%, but also reduces runtime and decreases token consumption. PMC, while achieving a moderate final pass rate of 30.00\%, requires a substantially longer runtime and higher token usage, reflecting inefficiencies in constraint handling. The sequential variant BF-Seq, which removes parallel planning and executes behavior trees sequentially, requires 428.25 s, confirming that the speedup of Behavior Forest primarily stems from its parallel execution design. This comparison highlights the critical role of parallel planning in accelerating reasoning without sacrificing solution quality.

\textit{\textbf{Summary}} With its parallel design, Behavior Forest achieves the best overall balance among the final pass rate, runtime, and token consumption. It delivers the highest final pass rate of 91.67\% while completing in only 122.13 s with 3210/1586 tokens. In contrast, the sequential variant BF-Seq requires 428.25 s, clearly demonstrating that the parallel architecture accelerates the reasoning process.

\section{Limitations and Future Work}
\textbf{Limitations}
Despite its effectiveness, Behavior Forest has several limitations. First, the framework relies on underlying databases as its information source. As a result, mechanisms for detecting unsafe or incorrect information are lacking. This limitation may lead to the generation of potentially risky plans when unreliable data are present. Second, our method is evaluated in two travel planning scenarios. Although these tasks are representative, the framework has not yet been validated in other multi-constraint domains, such as meeting planning, where constraints related to time, budget, and participant availability must be simultaneously considered. Finally, while our experiments demonstrate practical feasibility, the framework has not yet been deployed or evaluated in real-world data settings.

\textbf{Future Work}
Our work offers several promising directions for future research. First, the current coordination mechanism of Behavior Forest is primarily based on budget-based interactions among behavior trees. Future work will explore more flexible coordination strategies that enable multiple rounds of negotiation and trade-off reasoning, allowing the system to better optimize global plan quality and user satisfaction. Second, owing to its integration of backtracking, Behavior Forest is well-positioned for extension to real-world applications, such as meeting scheduling. Future work will evaluate the framework in these domains. Finally, future research may investigate how Behavior Forest can be used to equip end-to-end agents with structured reasoning capabilities, enabling scalable and systematic decision-making in complex environments.

\section{Conclusion}
In this work, we propose Behavior Forest, the core idea of which is to use a forest architecture to decouple complex planning tasks into parallel behavior trees, with each tree managing a subset of interdependent constraints. The forest architecture allows the LLM to reason concurrently across dimensions, while a global coordination mechanism ensures coordination among trees. Behavior Forest is evaluated on two benchmarks, TravelPlanner and ChinaTravel. It achieves a final pass rate of 94.44\% on the TravelPlanner dataset under the sole-planning setting, representing an improvement over the best existing method, llm-rwplanning (85\%). On the ChinaTravel dataset, it outperforms NeSy Planning by 27.28\% at the human level. 

\bibliographystyle{IEEEtran} 
\bibliography{ref} 

\begin{IEEEbiography}
[{\includegraphics[width=0.75in,height=0.9in,clip,keepaspectratio]{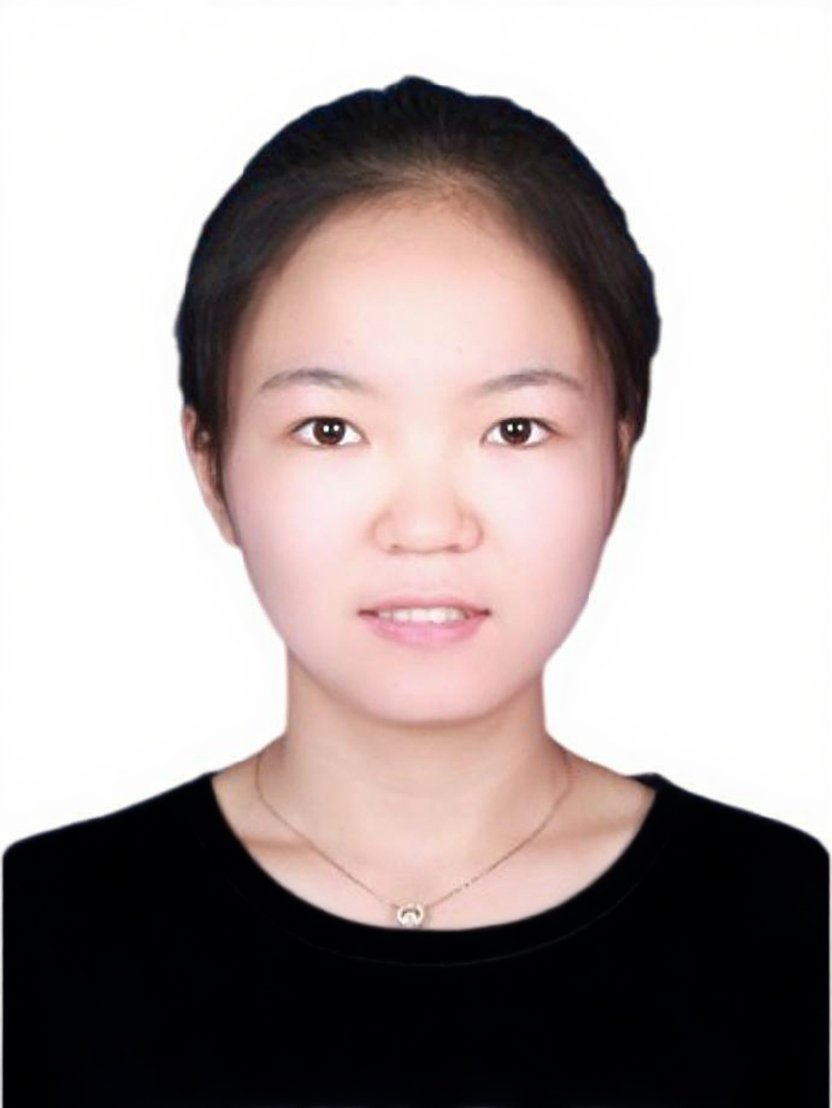}}]{Duanyang Yuan} is currently pursuing a Ph.D. degree at the National University of Defense Technology (NUDT). Her research interests include knowledge graphs, large language models, information retrieval, and retrieval-augmented generation. She has published several papers in AAAI, Frontiers in Physiology, etc.
\end{IEEEbiography}

\begin{IEEEbiography}
[{\includegraphics[width=0.75in,height=0.9in,clip,keepaspectratio]{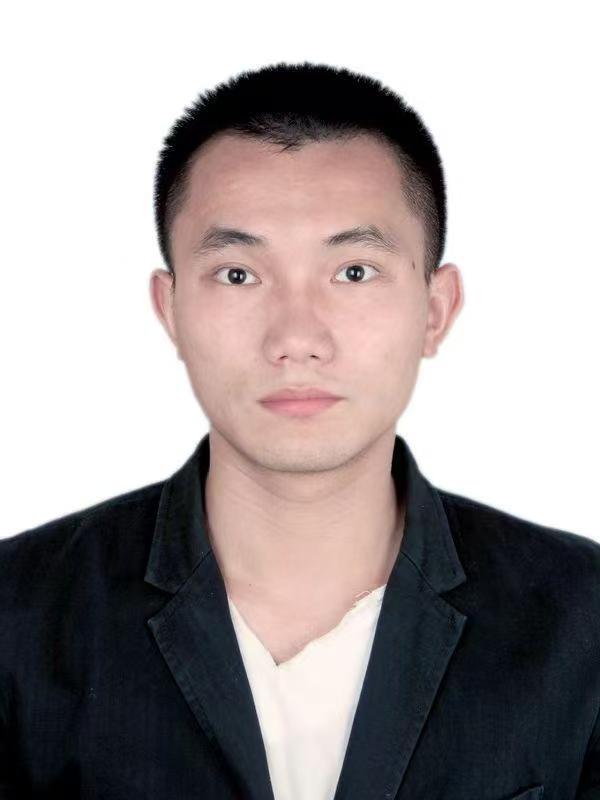}}]{Sihang Zhou} received his PhD degree from School of Computer, National University of Defense Technology (NUDT), China. He is now an associate professor at the College of Intelligence Science and Technology, NUDT. His current research interests include machine learning, knowledge graph, and medical image analysis. Dr. Zhou has published 40+ peer-reviewed papers, including IEEE T-IP, IEEE T-NNLS, IEEE T-MI, Medical Image Analysis, AAAI, MICCAI, etc.
\end{IEEEbiography}

\begin{IEEEbiography}
[{\includegraphics[width=0.75in,height=0.9in,clip,keepaspectratio]{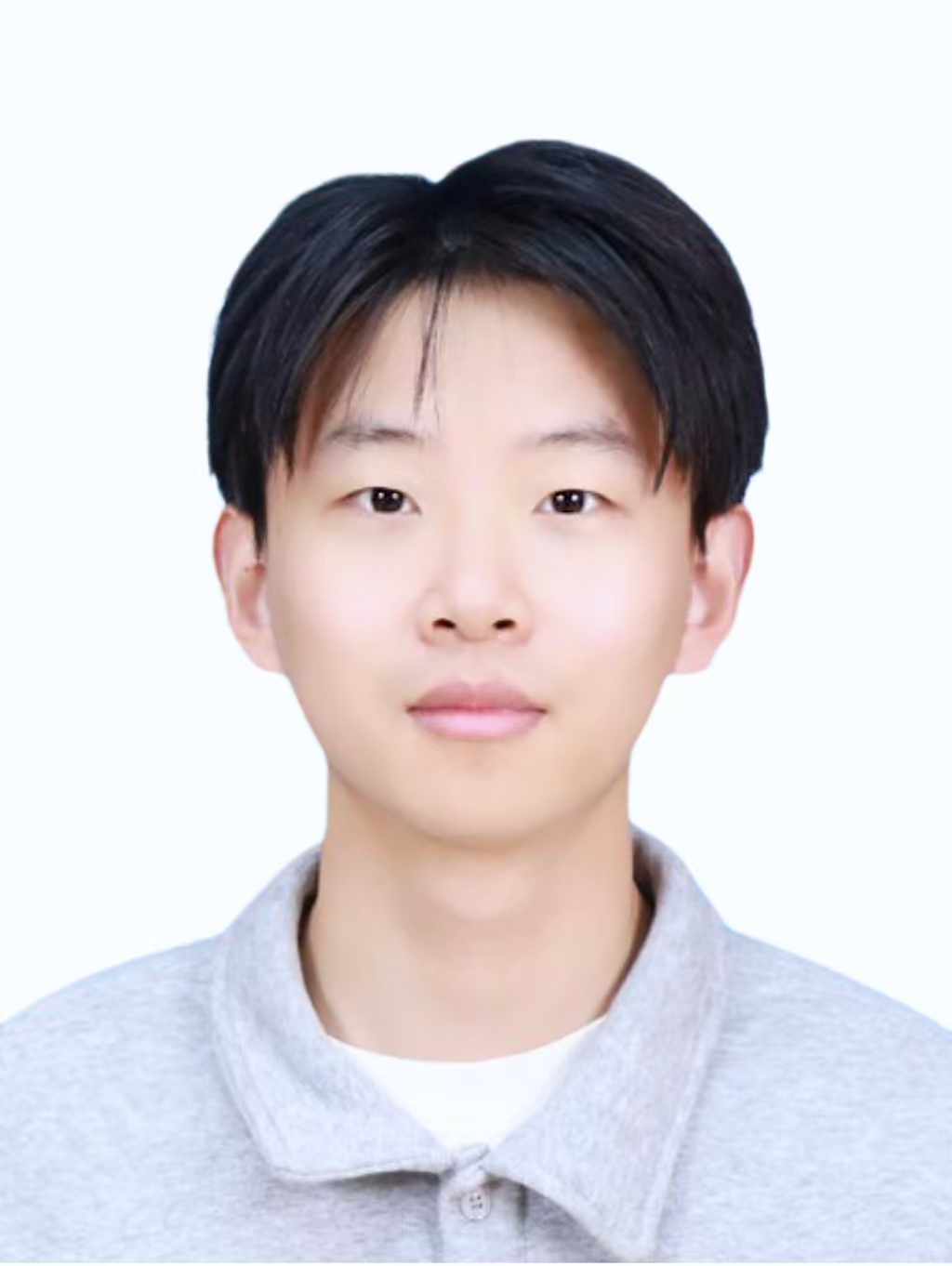}}]{Yanning Hou} is currently pursuing a Master’s degree at the National University of Defense Technology (NUDT). His research interests include knowledge graphs, large language models, retrieval-augmented generation, and multimodal learning. He has published papers in international conferences such as IJCAI and PRCV, focusing on intelligent reasoning and knowledge-enhanced representation learning.
\end{IEEEbiography}

\begin{IEEEbiography}[{\includegraphics[width=0.75in,height=0.9in,clip,keepaspectratio]{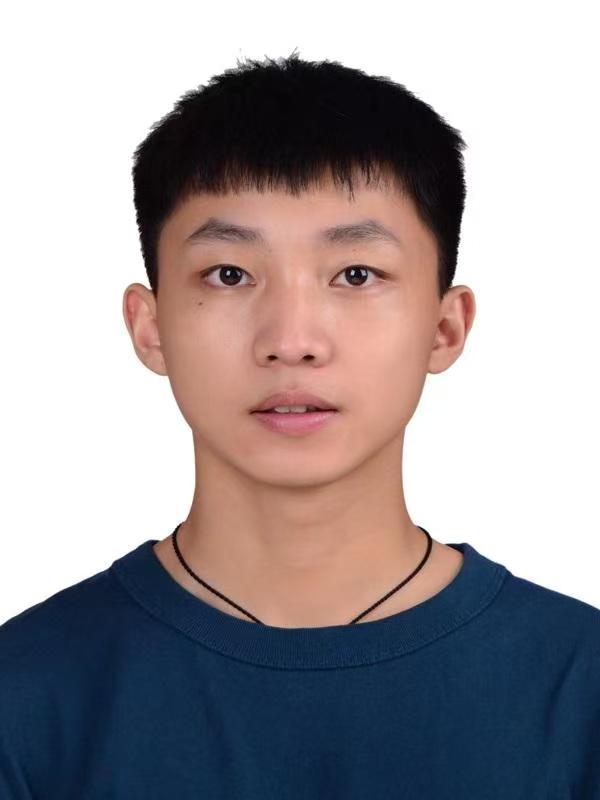}}] {Xiaoshu Chen} is currently pursuing a Ph.D. degree at the National University of Defense Technology (NUDT). Before joining NUDT, he got his MSc degree at WuHan University (WHU) and BSc degree at China University of Mining and Technology (CUMT). His research interests include large language models, knowledge graph, information retrieval and semantic segmentation. He has published several papers in AAAI, IEEE TGRS and Neurocomputing, etc.
\end{IEEEbiography}

\begin{IEEEbiography}[{\includegraphics[width=0.75in,height=0.9in,clip,keepaspectratio]{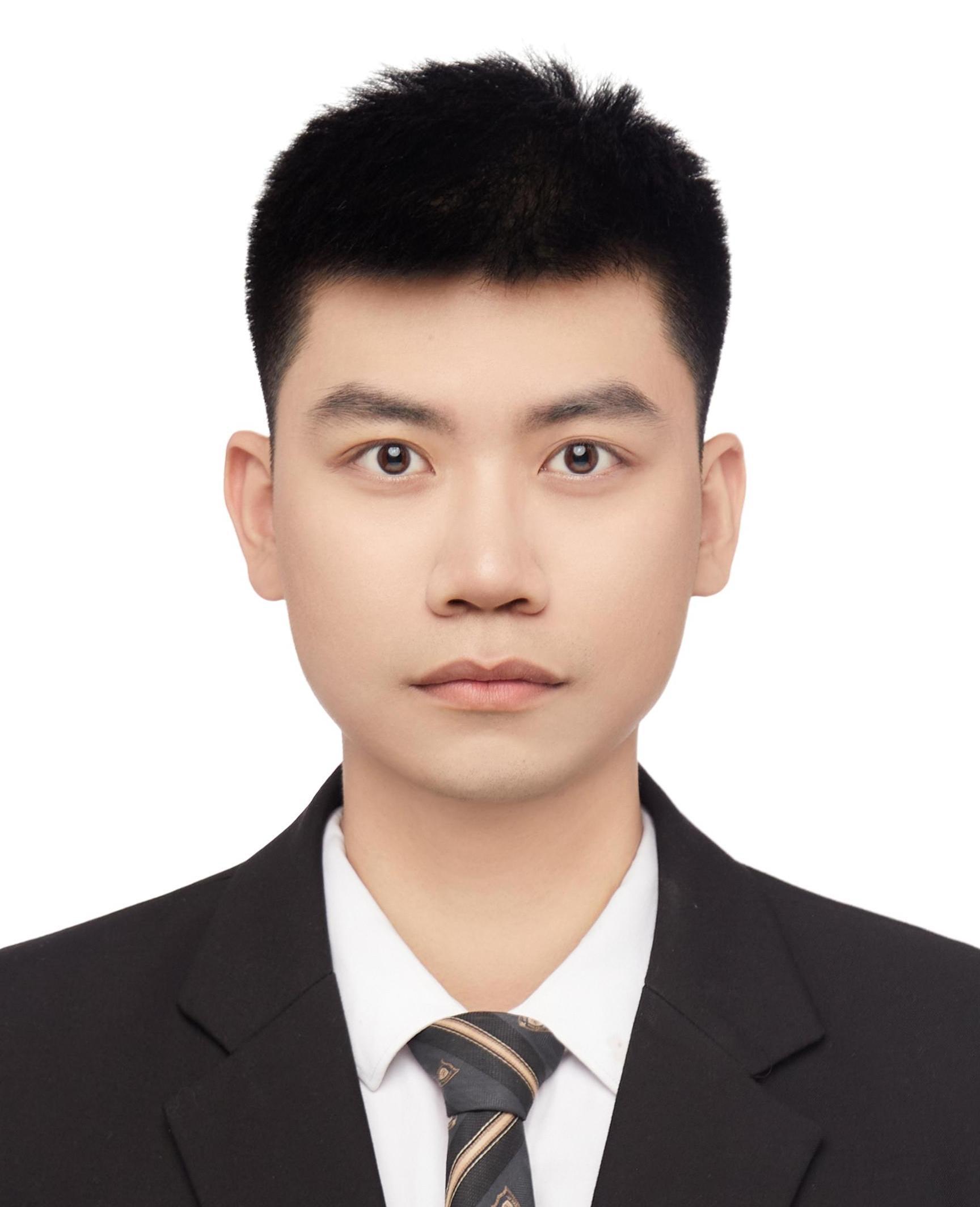}}]
{Haoyuan Chen} is currently pursuing a Ph.D. degree at the National University of Defense Technology (NUDT). His research interests include large language models, retrieval-augmented generation, and multi-agent systems.
\end{IEEEbiography}

\begin{IEEEbiography}[{\includegraphics[width=0.75in,height=0.9in,clip,keepaspectratio]{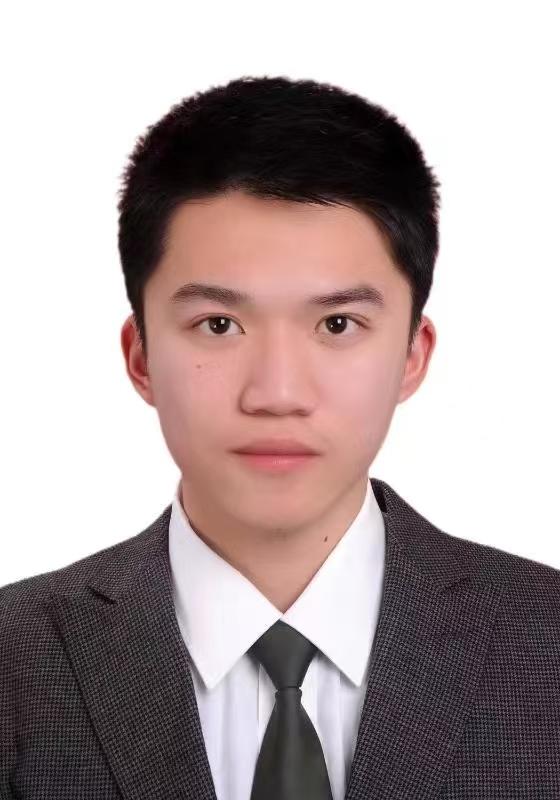}}]
{Ke Liang} received his Ph.D. degree at the National University of Defense Technology (NUDT). He got his BSc degree at Beihang University (BUAA) and received his MSc degree from the Pennsylvania State University (PSU). His current research interests include knowledge graphs, graph learning, and healthcare AI. He has published several papers in highly regarded journals and conferences such as SIGIR, AAAI, ICML, ACM MM, IEEE TNNLS, IEEE TKDE, etc.
\end{IEEEbiography}

\begin{IEEEbiography}
[{\includegraphics[width=0.75in,height=0.9in,clip,keepaspectratio]{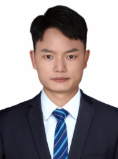}}] {Jiyuan Liu} is a lecturer with the College of Systems Engineering, National University of Defense Technology (NUDT), China. He received the Ph.D. degree from the NUDT in 2022. Previously, he was an undergraduate of QianxueSen Class (QXSC) at NUDT from 2013 to 2017, an visiting student of Jiangchuan Liu's lab with the support from China Scholarship Council (CSC) from 2016 to 2017.  His research interest is theory and algorithms of statistical machine learning and its applications, especially in multi-view learning, federated learning, anomaly detection and contrastive learning.  He has published several papers in highly regarded journals and conferences such as TPAMI, TCSVT, ICML, AAAI, NeurIPS, IEEE TNNLS, etc.
\end{IEEEbiography}

\begin{IEEEbiography}
[{\includegraphics[width=0.75in,height=0.9in,clip,keepaspectratio]{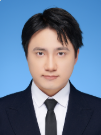}}] {Chuan Ma} is an Associate Professor with the School of Computer Science, Chongqing University, China, and serves as the Secretary-General of the IEEE International Standard Committee on Trusted Data Sharing. He received the B.Sc. degree from Beijing University of Posts and Telecommunications in 2013 and the Ph.D. degree from the University of Sydney in 2018. His research interests lie in machine learning, stochastic optimization, and artificial intelligence. He has published over 40 papers in related fields, including more than 20 SCI-indexed journal papers and 2 ESI highly cited papers, with publications in top venues such as Nature Machine Intelligence. 
\end{IEEEbiography}

\begin{IEEEbiography}
[{\includegraphics[width=0.75in,height=0.9in,clip,keepaspectratio]{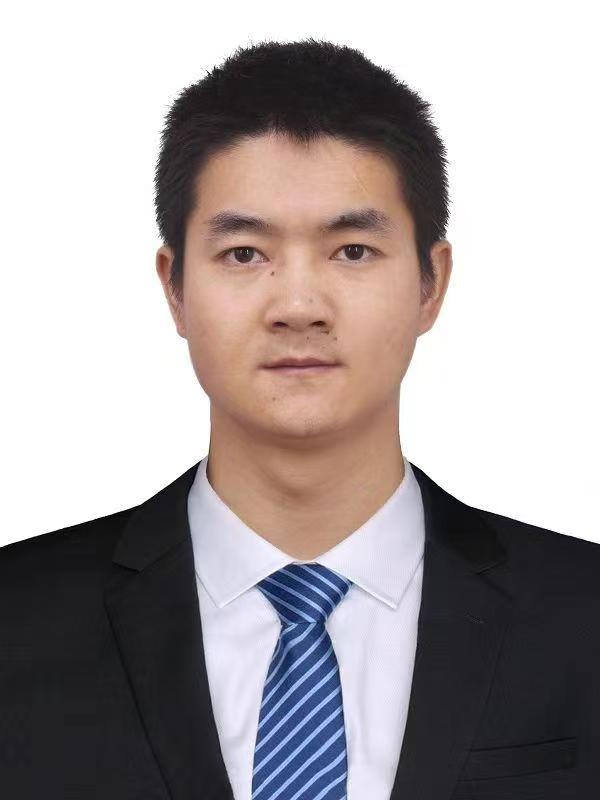}}] {Xinwang Liu} received his PhD degree from National University of Defense Technology (NUDT), China. He is now Professor of School of Computer, NUDT. His current research interests include kernel learning and unsupervised feature learning. Dr. Liu has published 60+ peer-reviewed papers, including those in highly regarded journals and conferences such as IEEE T-PAMI, IEEE T-KDE, IEEE T-IP, IEEE T-NNLS, IEEE T-MM, IEEE T-IFS, ICML, NeurIPS, ICCV, CVPR, AAAI, IJCAI, etc. He serves as the associated editor of T-NNLS, T-CYB and Information Fusion Journal. More information can be found at {https://xinwangliu.github.io/}.
\end{IEEEbiography}

\begin{IEEEbiography}
[{\includegraphics[width=0.75in,height=0.9in,clip,keepaspectratio]{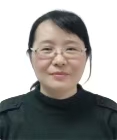}}] 
{Jian Huang} received the Ph.D. degree from the College of Mechatronics and Automation, National University of Defense Technology, Changsha, China, in 2000. She is now a professor at the College of Intelligence Science, National University of Defense Technology, China. Her current research interests include system simulation and artificial intelligence.
\end{IEEEbiography}

\vfill

\end{document}